\documentclass[11pt, a4paper, onecolumn, confidential, copyright, goog]{google}
\usepackage[utf8]{inputenc}
\usepackage{longtable}
\usepackage{booktabs}
\usepackage{tabularx}
\usepackage{ragged2e}  
\usepackage{amsmath}
\usepackage{enumitem}
\usepackage{amssymb} 
\usepackage{amsthm}
\usepackage{graphicx}
\usepackage{microtype}
\usepackage{adjustbox}
\usepackage{multicol}
\usepackage{multirow}
\usepackage{colortbl} 
\usepackage{pifont}    
\usepackage[most]{tcolorbox}
\usepackage{wrapfig}
\usepackage[table]{xcolor}
\usepackage[inkscapelatex=false]{svg}

\usepackage[utf8]{inputenc} 
\usepackage[T1]{fontenc}    
\usepackage{hyperref}       
\usepackage{url}            
\usepackage{amsfonts}       
\usepackage{nicefrac}       
\usepackage{xcolor}         
\usepackage[utf8]{inputenc}
\usepackage[T1]{fontenc}
\usepackage{hyperref}
\usepackage{url}
\usepackage{booktabs}
\usepackage{amsmath}
\usepackage{amsfonts}
\usepackage{nicefrac}
\usepackage{microtype}
\usepackage{graphicx}
\usepackage[numbers, square, sort&compress]{natbib}
\usepackage{cleveref}
\usepackage{xcolor}
\usepackage{tikz}
\usetikzlibrary{arrows.meta,positioning,fit,shapes.geometric,calc}
\usepackage{enumitem}
\usepackage{listings}
\usepackage{titletoc}
\usepackage{hyperref}
\usepackage{adjustbox}

\definecolor{jevblue}{HTML}{1F5FA8}
\definecolor{jevgrey}{HTML}{8A8A80}
\newcommand{\resMedqaJevAcc}{72.5}
\newcommand{\resMedqaJevAccCI}{69.8--75.3}

\newcommand{\resMedqaJevEce}{0.039}

\newcommand{\resMedqaBaseDelta}{-3.9}
\newcommand{\resMedqaBaseDeltaCI}{-6.3 to -1.4}
\newcommand{\resMedqaMedgemmaAcc}{47.7}

\newcommand{\resMedqaMedgemmaDelta}{-24.8}

\newcommand{\resMedqaGemmaAcc}{39.7}

\newcommand{\resMedqaSmallAcc}{39.4}

\newcommand{\resMedqaPythiaAcc}{19.2}

\newcommand{\resUsmleJevAcc}{75.4}
\newcommand{\resUsmleJevAccCI}{73.0--77.6}

\newcommand{\resUsmleJevEce}{0.028}

\newcommand{\resUsmlePlattEce}{0.038}

\newcommand{\resUsmleBaseDelta}{-4.6}
\newcommand{\resUsmleBaseDeltaCI}{-6.8 to -2.4}
\newcommand{\resUsmleMedgemmaAcc}{54.0}

\newcommand{\resUsmleGemmaAcc}{47.1}

\newcommand{\resUsmleGemmaEce}{0.515}

\newcommand{\resUsmleSmallAcc}{43.4}

\newcommand{\resUsmlePythiaAcc}{27.4}

\newcommand{\resNejmJevAcc}{80.1}
\newcommand{\resNejmJevAccCI}{76.6--83.4}

\newcommand{\resNejmJevEce}{0.063}

\newcommand{\resNejmPlattEce}{0.061}

\newcommand{\resNejmBaseDelta}{-3.1}

\newcommand{\resNejmMedgemmaAcc}{60.4}

\newcommand{\resNejmMedgemmaEce}{0.210}

\newcommand{\resNejmGemmaAcc}{40.6}

\newcommand{\resNejmSmallAcc}{48.8}

\newcommand{\resNejmPythiaAcc}{20.7}

\newcommand{\resDiagarenaJevAcc}{55.5}
\newcommand{\resDiagarenaJevAccCI}{52.5--58.8}

\newcommand{\resDiagarenaJevEce}{0.071}

\newcommand{\resDiagarenaPlattEce}{0.112}

\newcommand{\resDiagarenaBaseDelta}{-1.1}

\newcommand{\resDiagarenaMedgemmaAcc}{42.3}

\newcommand{\resDiagarenaMedgemmaEce}{0.376}

\newcommand{\resDiagarenaMedgemmaDelta}{-13.2}

\newcommand{\resDiagarenaGemmaAcc}{36.3}

\newcommand{\resDiagarenaGemmaEce}{0.626}

\newcommand{\resDiagarenaSmallAcc}{41.7}

\newcommand{\resDiagarenaPythiaAcc}{23.3}

\newcommand{\resTriageDispPythiaAcc}{4.2}

\newcommand{\resTriageDispXgbDelta}{+8.0}
\newcommand{\resTriageDispXgbDeltaCI}{+4.9 to +11.1}

\newcommand{\resTriageJevAuroc}{0.745}

\newcommand{\resTriageJevEce}{0.147}

\newcommand{\resTriagePlattEce}{0.041}

\newcommand{\resTriagePythiaAuroc}{0.490}

\newcommand{\resTriageXgbAuroc}{0.809}

\newcommand{\resTriageXgbEce}{0.042}

\newcommand{\resTriageXgbDelta}{+0.064}
\newcommand{\resTriageXgbDeltaCI}{+0.034 to +0.093}

\newcommand{\resEdmedsBaseDelta}{+0.026}

\newcommand{\resEdmedsXgbDelta}{+0.081}
\newcommand{\resEdmedsXgbDeltaCI}{+0.053 to +0.113}

\newcommand{\resMortalityJevAuroc}{0.926}
\newcommand{\resMortalityJevAurocCI}{0.881--0.960}

\newcommand{\resMortalityJevMeanProb}{0.312}
\newcommand{\resMortalityJevEce}{0.286}

\newcommand{\resMortalityPlattEce}{0.007}

\newcommand{\resMortalityBaseEce}{0.426}

\newcommand{\resMortalityBaseDelta}{-0.027}

\newcommand{\resMortalitySmallAuroc}{0.711}

\newcommand{\resMortalityPythiaAuroc}{0.764}

\newcommand{\resMortalityXgbAuroc}{0.902}

\newcommand{\resMortalityXgbEce}{0.014}

\newcommand{\resMortalityXgbDelta}{-0.023}
\newcommand{\resMortalityXgbDeltaCI}{-0.054 to +0.006}
\newcommand{\resMortalityTfidfAuroc}{0.879}

\newcommand{\resMortalityTfidfDelta}{-0.046}
\newcommand{\resMortalityTfidfDeltaCI}{-0.089 to -0.007}

\newcommand{\resMortalityPrev}{2.6}

\newcommand{\resLongLOSJevEce}{0.191}

\newcommand{\resLongLOSPlattEce}{0.027}

\newcommand{\resLongLOSBaseEce}{0.348}

\newcommand{\resLongLOSBaseDelta}{-0.024}

\newcommand{\resLongLOSSmallAuroc}{0.421}

\newcommand{\resLongLOSPythiaAuroc}{0.629}

\newcommand{\resLongLOSXgbEce}{0.059}

\newcommand{\resLongLOSTfidfDelta}{+0.045}
\newcommand{\resLongLOSTfidfDeltaCI}{+0.011 to +0.080}

\newcommand{\resReadmissionJevEce}{0.272}

\newcommand{\resReadmissionPlattEce}{0.024}

\newcommand{\resReadmissionBaseEce}{0.372}

\newcommand{\resReadmissionSmallAuroc}{0.501}

\newcommand{\resReadmissionXgbEce}{0.046}

\newcommand{\resReadmissionXgbDelta}{+0.012}
\newcommand{\resReadmissionXgbDeltaCI}{-0.028 to +0.053}

\newcommand{\resNejmGptThreeMin}{37.7}
\newcommand{\resNejmGptThreeMax}{47.5}
\newcommand{\resNejmGptFourMin}{70.3}
\newcommand{\resNejmGptFourMax}{75.8}
\newcommand{\resMedqaRefGptFour}{74.7}
\newcommand{\resUsmleRefGptFour}{78.9}

\newcommand{\resMedqaRefGptThree}{44.6}
\newcommand{\resUsmleRefGptThree}{50.8}

\newcommand{\resDiagarenaRefGptFourO}{36.4}
\newcommand{\resDiagarenaRefOone}{61.9}

\newcolumntype{L}{>{\raggedright\arraybackslash}X}
\newcolumntype{P}[1]{>{\raggedright\arraybackslash}p{#1}}
\usepackage[most]{tcolorbox} 

\uselogo{} 

\title{MedJev: Evaluating Zero-Shot System-One Decision Making in Medicine}

\correspondingauthor{simonlee711@g.ucla.edu}

\author[1]{Simon A. Lee}
\author[1]{Jiahang Sha}
\author[1]{Moonseong Jeong}
\author[1]{Yunbei Pan}
\author[1,2]{Jeffrey N. Chiang}

\affil[1]{UCLA Computational Medicine}
\affil[2]{UCLA Neurosurgery}

\begin{abstract}
System-One decision models such as Jev answer structured questions about a given state, including option selection and probability estimation, in a single forward pass without generating text. Their speed, low computational cost, and ability to produce calibrated probabilities make them promising for clinical decision support. We investigate the zero-shot performance of MedJev, an open System-One model that runs locally on a laptop, across ten medical benchmarks. Four benchmarks assess medical knowledge and diagnosis (MedQA, USMLE, NEJM, and DiagnosisArena), while six evaluate patient-level decisions using cohorts from MIMIC-IV (In-Hospital Mortality, Length of Stay $>$ 7 Days, 30-Day Readmission, and ED Admission at 2\,h) and MIMIC-IV-ED (ED Triage Admission and ED Triage Disposition). We compare MedJev against zero-shot LLM baselines, including Qwen3.5-9B, MedGemma-1.5-4B, Gemma-3-4B, Qwen3.5-2B, and Pythia-2.8B. We also compare against published results for GPT-3.5, GPT-4, GPT-4o, and o1 on the same benchmarks, as well as XGBoost and TF-IDF models. In the zero-shot setting, MedJev achieves accuracies of \resMedqaJevAcc\% on MedQA, \resUsmleJevAcc\% on USMLE, and \resNejmJevAcc\% on NEJM, with well-calibrated confidence estimates (ECE \resUsmleJevEce--\resNejmJevEce). Its accuracy is within 2--4 percentage points of published zero-shot GPT-4 results on MedQA and USMLE (\resMedqaRefGptFour\% and \resUsmleRefGptFour\%, respectively) and substantially exceeds that of GPT-3.5 (\resMedqaRefGptThree\% and \resUsmleRefGptThree\%). On patient-level tasks, MedJev achieves an AUROC of \resMortalityJevAuroc{} for In-Hospital Mortality, matching or exceeding supervised models (XGBoost \resMortalityXgbAuroc), and performs comparably on 30-Day Readmission. However, it underperforms supervised models on operational outcomes such as ED Triage Admission (\resTriageJevAuroc{} vs.\ \resTriageXgbAuroc). Moreover, its patient-level probability estimates do not account for local outcome prevalence (In-Hospital Mortality ECE \resMortalityJevEce). To address this limitation, we introduce a lightweight, two-parameter recalibration method that adapts predicted probabilities to local patient populations using 50--200 labeled examples, reducing the In-Hospital Mortality ECE to \resMortalityPlattEce{}. These findings highlight the potential of open System-One models as a new baseline for medical AI, offering a competitive zero-shot alternative to general-purpose LLMs and EHR-specific predictive models. We release our code and evaluation harness to support future research.\\ \\
Github: \href{https://github.com/Simonlee711/MedJev}{https://github.com/Simonlee711/MedJev}\\
HuggingFace: \href{https://huggingface.co/Simonlee711/MedJev}{https://huggingface.co/Simonlee711/MedJev}

\end{abstract}

\begin{document}

\maketitle

\section{Introduction}

Large language models (LLMs) are increasingly evaluated on medical tasks, including clinical question answering, diagnosis, and patient outcome prediction \cite{singhal2023medpalm}. Most evaluations use generative models that produce textual answers, which are subsequently parsed into decisions. However, many medical prediction tasks require a structured output rather than a generated response. For example, predicting in-hospital mortality requires estimating the probability of death given the information available at a particular point in a patient's hospitalization, while emergency department triage requires selecting among possible dispositions. These tasks can be expressed as decisions over a fixed set of outcomes, making them suitable for models that directly estimate probabilities without generating text.

System-One models provide an alternative to generative inference for such tasks. Inspired by the distinction between fast and deliberative reasoning \cite{kahneman2011thinking}, System-One models directly answer structured questions about a given state in a single forward pass. Jev \cite{typesafe2026jev} implements this approach through typed requests that pair an input state, represented as text or JSON, with questions specifying the desired output. A \texttt{CHOICE} request returns a probability distribution over candidate answers, while a \texttt{NOUL} request estimates the probability of a binary outcome. Unlike conventional generative evaluation, which relies on decoding and parsing model responses, this interface produces structured probabilities directly. It therefore provides a common representation for evaluating multiple-choice medical questions and patient-level predictions from electronic health records (EHRs).

Despite this interface, the applicability of System-One models to medical decision tasks remains unclear. Previous evaluations of Jev have demonstrated competitive zero-shot performance across general-purpose benchmarks \cite{deusser2026jev}, but medical knowledge, diagnostic reasoning, and EHR-based prediction present different challenges. Medical knowledge benchmarks require selecting correct answers from clinical vignettes that mimic differential diagnosis, whereas patient-level prediction requires interpreting heterogeneous longitudinal records and estimating outcomes whose prevalence may vary across clinical populations. Furthermore, a model that accurately ranks patients by risk does not necessarily produce probabilities that reflect observed outcome frequencies. Evaluating System-One models in medicine therefore requires examining both their predictive performance and probability calibration across different task types.

In this work, we study how an open System-One model can be incorporated into a medical evaluation pipeline. We release MedJev which builds on OpenJev \cite{openjev2026flash} and run a suite of evaluations. Our evaluation represents each benchmark as a Jev request consisting of an input state and a typed question. Medical question-answering and diagnosis benchmarks are formulated as \texttt{CHOICE} requests, while patient-level outcomes are primarily formulated as \texttt{NOUL} requests. For EHR-based tasks, we construct patient states using only information available at a predefined prediction time. This formulation allows the same inference interface to be applied across medical knowledge questions, longitudinal patient records, and emergency department triage data.

We evaluate MedJev on ten medical benchmarks (expanding on the work of \cite{madridgarcía2026jev}) spanning medical knowledge, differential diagnosis, and patient-specific prediction. These include MedQA \cite{jin2021disease}, USMLE \cite{jin2021disease}, NEJM Case Challenges, and DiagnosisArena \cite{zhu2026diagnosisarena}, together with six outcomes derived from MIMIC-IV \cite{johnson2023mimic} and MIMIC-IV-ED \cite{johnson2023mimic}. To assess the effect of System-One training, we compare MedJev against its underlying language model and other open-weight LLMs using identical prompts and probability readouts. We additionally consider published results from frontier LLMs and train supervised XGBoost and TF-IDF baselines for EHR specific tasks. Beyond predictive accuracy and discrimination, we evaluate probability calibration and examine whether lightweight post-hoc recalibration can adapt zero-shot predictions to local outcome prevalence. We also vary the number of labeled examples available to supervised models to characterize when task-specific training becomes preferable to zero-shot inference.

Our results show that System-One models can serve as competitive zero-shot predictors across several medical tasks, although their effectiveness depends on the prediction target. MedJev achieves strong performance on medical knowledge benchmarks and competitive risk discrimination for outcomes such as in-hospital mortality, while supervised models perform better on operational outcomes such as emergency department admission. Although its confidence estimates are well calibrated on medical question-answering benchmarks, its patient-level probabilities require adjustment to local outcome frequencies. A lightweight recalibration procedure using a small number of labeled examples substantially reduces this error without changing patient risk rankings. These findings illustrate how System-One models can be evaluated within a unified medical prediction pipeline and clarify the role of zero-shot inference, local calibration, and supervised learning in clinical prediction settings.

\section{Related Work}
\label{sec:related}

\paragraph{System-One decision models.}
Most language models perform decision-making tasks through autoregressive generation, producing textual responses that must be interpreted or parsed to recover a final prediction. System-One decision models instead formulate inference as a direct mapping from an observed state and a structured question to a probability distribution over possible outcomes. Jev \citep{typesafe2026jev} implements this paradigm through three typed primitives: \textsc{Choice}, which returns a distribution over a finite set of alternatives; \textsc{Noul}, which estimates the probability of a binary outcome; and \textsc{Score}, which represents judgments on an ordinal scale. These primitives allow classification, probability estimation, and ordinal assessment to share a common inference interface without requiring autoregressive decoding. Deu{\ss}er et. al, \cite{deusser2026jev} systematically evaluate Jev in the zero-shot setting across 37 public datasets, comparing its predictive performance and computational efficiency against open-weight LLMs evaluated using exact option probabilities. Their results establish the feasibility of System-One inference for general-purpose classification and decision tasks. However, their evaluation includes only one biomedical dataset, PubMedQA \citep{jin2019pubmedqa}, and does not examine patient-level prediction from EHRs. More recently, OpenJev \citep{openjev2026flash} provides an open-weight implementation of the Jev interface by fine-tuning Qwen3.5-9B \citep{qwen2026qwen35}. OpenJev obtains structured predictions from the probability distribution over option-letter tokens, followed by temperature calibration. Our study extends prior System-One evaluations to medical knowledge, diagnosis, and longitudinal patient prediction, while examining both zero-shot predictive performance and the reliability of the resulting probabilities.

\paragraph{Medical question answering and diagnosis.}
Medical question answering has become a standard approach for evaluating the clinical knowledge and reasoning capabilities of language models \cite{kim2026ehrnote, lee2024can, li2024mediq, lin2025case, lee2025modern}. MedQA \citep{jin2021medqa} evaluates medical reasoning using multiple-choice questions derived from professional medical examinations, including the United States Medical Licensing Examination (USMLE). MedMCQA \citep{pal2022medmcqa} provides a large collection of multiple-choice questions spanning medical subjects, while PubMedQA \citep{jin2019pubmedqa} evaluates biomedical question answering using research abstracts. These benchmarks have been widely used to measure the medical knowledge encoded in pretrained and instruction-tuned language models. Work on medical LLMs, including Med-PaLM \citep{singhal2023medpalm}, demonstrates that large generative models can achieve strong performance on medical examination questions, motivating their use as reference models for medical AI evaluation. However, performance on examination-style questions does not necessarily characterize a model's ability to interpret complex clinical presentations or make patient-specific predictions.

Diagnostic benchmarks address this limitation by evaluating models on clinical cases rather than isolated knowledge questions. DiagnosisArena \citep{zhu2026diagnosisarena} constructs diagnostic tasks from published case reports, requiring models to distinguish among candidate diagnoses using clinical histories, examinations, and investigation findings. \cite{savage2024diagnostic} investigate the diagnostic performance of GPT-3.5 and GPT-4 on clinical vignettes using different reasoning prompts and clinician-graded open-ended responses. These studies predominantly evaluate generative models through textual answers, often allowing reasoning or explanation before a final diagnosis. In contrast, our evaluation formulates medical knowledge and diagnostic tasks as structured multiple-choice decisions, enabling direct probability-based comparisons among System-One models and conventional LLMs. Because multiple-choice and open-ended diagnosis differ in difficulty, previously reported generative results are treated as external reference points rather than strictly equivalent evaluations.

\paragraph{Clinical prediction from electronic health records.}
Clinical prediction from multimodal clinical data \cite{an2025raptor, zhou2026physiology, huerta2026multi, lee2026multimodal, pan2026hipno, lee2025towards,  larey2026jepa, lee2026himae} and EHRs \cite{steinberg2024motor, jing2026one, wornow2023shaky} differs from medical question answering because models must reason over heterogeneous patient information collected across time. Established prediction tasks include in-hospital mortality, length of stay, and hospital readmission, which have been widely studied using critical care datasets such as MIMIC \citep{johnson2023mimiciv,harutyunyan2019benchmark, kolo2024meds, jeanselme2025fomoh}. These tasks require predictions at clinically meaningful decision times, making the availability and temporal ordering of patient information important aspects of benchmark construction. The Medical Event Data Standard (MEDS) \citep{arnrich2024medical, mcdermott2025meds, mcdermott2026meds} provides a standardized representation of longitudinal patient records as timestamped clinical events, facilitating the development and evaluation of models across EHR-derived tasks \cite{steinberg2024motor, lee2026forecasting, jing2026one, an2025dk}.

Emergency department prediction presents a related but distinct setting in which decisions must be made using information available early in the patient encounter. \cite{hong2018triage} study hospital admission prediction at emergency department triage using routinely collected patient characteristics, vital signs, and chief complaints. Gradient-boosted trees and other supervised tabular models provide strong baselines for such structured clinical prediction tasks. Language models offer an alternative by representing patient records as text and applying pretrained language understanding capabilities to the resulting sequences. \cite{lee2025clinical} investigate this approach using serialized emergency department records, while EHRSHOT \citep{wornow2023ehrshot} introduces a framework for evaluating foundation models on EHR prediction tasks under limited-label settings. These studies motivate the evaluation of pretrained models on patient-level outcomes, but their settings generally involve either supervised adaptation, learned patient representations, or conventional language-model inference. We instead examine whether a System-One model can directly estimate clinical outcome probabilities from serialized EHR states without task-specific training.

\paragraph{Language models for structured data and probability calibration.}
Applying language models to structured clinical records is closely related to research on language-based representations of tabular data \cite{lee2025clinical, pmlr-v281-lee25a, jaitly2023towards, fleming2024medalign}. TabLLM \citep{hegselmann2023tabllm} demonstrates that serializing tabular features into natural-language descriptions enables pretrained language models to perform classification with relatively few labeled examples. This approach is particularly relevant when task-specific training data are scarce, although conventional supervised methods often become more competitive as the number of labeled examples increases. \cite{grinsztajn2022tree} further demonstrate the strong performance of tree-based methods on typical tabular prediction tasks, motivating their inclusion as supervised baselines. For clinical prediction, this comparison is important because pretrained models may provide useful predictions without local labels, whereas models trained directly on EHR data can learn relationships specific to the target population and outcome.

Beyond discrimination and accuracy, clinical prediction requires an assessment of probability calibration. A model may correctly rank patients by risk while systematically overestimating or underestimating the frequency of an outcome. This distinction is particularly important when a model is evaluated zero-shot on a patient population whose outcome prevalence differs from the distributions encountered during pretraining. Post-hoc calibration methods provide a lightweight mechanism for correcting predicted probabilities without retraining the underlying model. Platt scaling \citep{platt1999probabilistic} learns a logistic transformation of model scores, while temperature scaling \citep{guo2017calibration} adjusts the sharpness of a predicted probability distribution. These approaches preserve binary risk rankings or multiclass predicted labels, respectively, while modifying the reported confidence. In our evaluation, we distinguish the zero-shot predictive capabilities of System-One models from the calibration of their probabilities on local patient populations. We investigate whether a small number of labeled examples is sufficient for post-hoc recalibration and compare this adaptation strategy against supervised models trained under the same label budgets.

\section{Methodology}
\label{sec:method}

We evaluate MedJev across ten medical benchmarks using a common pipeline that converts each task into a structured decision request. The pipeline consists of constructing an input state, specifying a question and its possible outcomes, obtaining a probability distribution from the model, and evaluating the prediction against the ground-truth outcome. All models are evaluated on the same test examples, with additional comparisons against supervised baselines and locally recalibrated predictions.

\subsection{System-One Inference}
\label{sec:jev}

A Jev request consists of a state $s$, represented as text or JSON, and a set of typed questions $Q$. We consider two question types. A \textsc{Choice} question specifies $K$ candidate outcomes and returns a probability distribution over them, while a \textsc{Noul} question estimates the probability of a binary outcome. This interface supports both medical multiple-choice questions and patient-level predictions without requiring autoregressive answer generation.

We use OpenJev Flash 9B \citep{openjev2026flash}, and adapt it on medical data (MIMIC-IV Note) to obtain MedJev throughout the evaluation. Each request is rendered as a prompt containing the state, question instructions, and candidate outcomes labeled with single-token letters. The model produces logits $\ell_1,\ldots,\ell_K$ for the candidate tokens at the first answer position. The corresponding probabilities are computed as
\begin{equation}
    p_i = \frac{\exp(\ell_i/T)}
    {\sum_{j=1}^{K}\exp(\ell_j/T)},
    \label{eq:readout}
\end{equation}
where $T$ is the readout temperature. For binary \textsc{Noul} questions, the two-option probability $p_1$ is further transformed as
\begin{equation}
    p_{\mathrm{yes}} =
    \sigma\left(
    \frac{1}{T_{\mathrm{N}}}
    \log\frac{p_1}{1-p_1}+b
    \right),
    \label{eq:noul}
\end{equation}
where $\sigma$ denotes the logistic sigmoid function. We use the published MLX configuration with $T=1.07$, $T_{\mathrm{N}}=1.0748$, and $b=0$, together with the frozen inference helper (SHA-256 prefix \texttt{81a22f1b}). The zero-shot LLM baselines use the same request construction and probability readout, with model-specific chat templates and tokenization where required.

\subsection{Medical Benchmarks}
\label{sec:tasks}

We evaluate four medical knowledge and diagnosis benchmarks and six patient-level prediction benchmarks. All tasks are formulated using the same Jev request interface, with \textsc{Choice} for categorical outcomes and \textsc{Noul} for binary outcomes. \Cref{tab:stats} summarizes the dataset sizes, prediction units, and label distributions.

\paragraph{Medical knowledge and diagnosis.}
We use MedQA, USMLE, NEJM, and DiagnosisArena to evaluate medical knowledge and diagnostic decision-making. MedQA and USMLE correspond to the five-option BigBio and four-option MedQA-USMLE releases of USMLE-style examination questions \citep{jin2021medqa}. DiagnosisArena \citep{zhu2026diagnosisarena} contains 915 publicly available case reports with four candidate diagnoses. NEJM \citep{savage2024diagnostic} contains 512 diagnostic vignettes originally evaluated using open-ended responses from GPT-3.5 and GPT-4.

All four benchmarks are represented as \textsc{Choice} requests. For NEJM, vignettes overlapping with MedQA retain their original five answer options. For the remaining cases, we construct five-option questions using the correct diagnosis, up to four clinician-graded incorrect GPT responses, and diagnoses from other vignettes when additional options are needed. Because this formulation differs from the original open-ended evaluation, published GPT results are used only as external references.

\paragraph{Longitudinal EHR prediction.}
We construct four patient-level benchmarks using MIMIC-IV \citep{johnson2023mimiciv} represented in the Medical Event Data Standard (MEDS) \citep{mcdermott2026meds}. The tasks are In-Hospital Mortality, Length of Stay $>$ 7 Days, 30-Day Readmission, and ED Admission at 2\,h. Each task is defined by an outcome and a fixed prediction time $t^\ast$.

In-Hospital Mortality and Length of Stay $>$ 7 Days are predicted 24\,h after hospital admission among adult patients who remain hospitalized at that time. 30-Day Readmission is predicted at discharge, and ED Admission at 2\,h is predicted two hours after emergency department arrival. For each prediction, the model receives only events within $(t^\ast-365\,\mathrm{d},t^\ast]$. The available events are serialized into a compact JSON state containing patient demographics, clinical history, care units, laboratory measurements, and medications when available.

We exclude information that would not be available at the specified prediction time. In particular, discharge-stamped diagnoses and diagnosis-related groups are excluded from the 24\,h prediction tasks. For ED Admission at 2\,h, we additionally remove admission-only demographic fields, ED and admission bookkeeping codes beginning 24\,h before arrival, and nonlaboratory events recorded after ED arrival. These restrictions prevent outcome-related information from entering the input state.

\paragraph{Emergency department triage.}
We construct two additional benchmarks from MIMIC-IV-ED \citep{johnson2023mimicived}: ED Triage Admission and ED Triage Disposition. Both are evaluated at the time of triage using information available at that point in the encounter. The input state includes age, sex, race, arrival transport, triage vital signs, pain score, Emergency Severity Index acuity, chief complaint, and emergency department visits within the preceding year. Subsequent medications, vital signs, diagnoses, and medication reconciliation records are excluded.

ED Triage Admission is a binary \textsc{Noul} task predicting whether the visit results in hospital admission. ED Triage Disposition is a six-option \textsc{Choice} task predicting discharge home, admission, transfer, departure before completion, death, or another disposition. \Cref{fig:state} provides an example of the serialized input and corresponding question.

\begin{figure}[t]
\begin{lstlisting}
{"decision_time": "at emergency department triage",
 "patient": {"age": 67, "sex": "male", "race": "WHITE"},
 "arrival_transport": "AMBULANCE",
 "triage": {
   "temperature_F": 100.9,
   "heart_rate_bpm": 118,
   "respiratory_rate": 24,
   "o2_saturation_pct": 91,
   "systolic_bp_mmHg": 96,
   "diastolic_bp_mmHg": 58,
   "pain": "4",
   "acuity_esi": 2,
   "chief_complaint": "Dyspnea"},
 "past_year_emergency_department_history": {
   "visits": 3,
   "visits_ending_in_admission": 2}}

Question (Noul):
Based on the information available at emergency
department triage, will this visit end with the
patient admitted to the hospital?
\end{lstlisting}
\caption{\textbf{Illustrative synthetic input for ED Triage Admission.} The state contains information available at triage and a binary question specifying the prediction target. No real patient information is shown.}
\label{fig:state}
\end{figure}

\paragraph{Dataset preparation.}
The longitudinal benchmarks use a MEDS-Transforms extract \citep{medstransforms} of MIMIC-IV v2.2 containing 299{,}712 patients distributed across five training shards, one tuning shard, and one held-out shard. Training examples are drawn from the training shards and test examples from the held-out shard. MIMIC-IV-ED v2.2 contains 425{,}087 emergency department stays. We retain adult patients with a corresponding MEDS birth record and assign each patient to the MEDS split associated with their \texttt{subject\_id}. This ensures patient-disjoint training and test partitions across the EHR benchmarks.

\begin{table}[t]
  \centering
  \caption{\textbf{Statistics of the ten medical benchmarks.}
  Train, validation, and test refer to the corresponding benchmark splits. EHR benchmarks use the MEDS \texttt{train}, \texttt{tuning}, and \texttt{held\_out} patient partitions. Patients denotes the number of distinct patients, and label balance denotes answer distributions or outcome prevalence.}
  \label{tab:stats}
  \small
  \resizebox{\linewidth}{!}{\begin{tabular}{lllrrrrl}
\toprule
Benchmark & Format & Unit & Train & Validation & Test & Patients & Label balance \\
\midrule
MedQA & \textsc{Choice} (5) & questions & 10{,}178 & 1{,}272 & 1{,}273 & -- & answers A--E: 21/22/20/21/16\% \\
USMLE & \textsc{Choice} (4) & questions & 10{,}178 & -- & 1{,}273 & -- & answers A--D: 28/24/27/21\% \\
NEJM & \textsc{Choice} (5) & vignettes & -- & -- & 512 & -- & answers A--E: 21/20/21/22/16\% \\
DiagnosisArena & \textsc{Choice} (4) & cases & -- & -- & 915 & -- & answers A--D: 24/25/24/27\% \\
\midrule
ED Triage Disposition & \textsc{Choice} (6) & ED stays & 339{,}994 & 42{,}463 & 42{,}625 & 205{,}500 & \shortstack[l]{home 56.8\%, admitted 37.2\%, left 3.2\%,\\transfer 1.7\%, other 1.0\%, expired 0.1\%} \\
ED Triage Admission & \textsc{Noul} & ED stays & 339{,}994 & 42{,}463 & 42{,}625 & 205{,}500 & 37.2\% positive \\
ED Admission at 2\,h & \textsc{Noul} & ED visits & 501{,}124 & 62{,}155 & 62{,}589 & 280{,}504 & 47.3\% positive \\
In-Hospital Mortality & \textsc{Noul} & admissions & 265{,}273 & 32{,}696 & 33{,}234 & 148{,}256 & 2.2\% positive \\
Length of Stay $>$ 7 Days & \textsc{Noul} & admissions & 265{,}273 & 32{,}696 & 33{,}234 & 148{,}256 & 21.5\% positive \\
30-Day Readmission & \textsc{Noul} & admissions & 338{,}477 & 41{,}988 & 42{,}181 & 176{,}934 & 20.2\% positive \\
\bottomrule
\end{tabular}
}
\end{table}

\subsection{Models and Baselines}
\label{sec:variants}

We compare zero-shot MedJev with conventional language models, previously reported frontier-model results, and supervised models trained on local EHR labels.

\paragraph{Zero-shot language models.}
We evaluate Qwen3.5-9B \citep{qwen2026qwen35}, the base model from which OpenJev was fine-tuned, to assess the effect of System-One training. Qwen3.5-2B provides a smaller model from the same family, while Pythia-2.8B \citep{biderman2023pythia} serves as a pretrained model without instruction tuning. On the medical knowledge and diagnosis benchmarks, we additionally evaluate MedGemma-1.5-4B \citep{sellergren2026medgemma} and its general-purpose base model, Gemma-3-4B \citep{gemmateam2025gemma3technicalreport}, to examine the effect of medical-domain specialization.

All models receive the same task states, question instructions, and candidate outcomes. Predictions are obtained from their option-token probabilities rather than generated answers. Each model uses its corresponding chat template when applicable. Pythia-2.8B receives the rendered prompt followed by \texttt{Answer:}; inputs exceeding its 2{,}048-token context window are truncated from the left to preserve the question and candidate options.

\paragraph{Published frontier-model references.}
We include previously reported GPT-3.5 and GPT-4 accuracies on the five-option MedQA and four-option USMLE benchmarks \citep{nori2023capabilities}. For NEJM, we report clinician-graded open-ended diagnostic accuracy from \cite{savage2024diagnostic}. We also include GPT-4o and o1 results reported on DiagnosisArena \citep{zhu2026diagnosisarena}. These models are not re-evaluated in our pipeline. Differences in prompting, answer format, and benchmark subsets preclude paired statistical comparisons. In particular, the published DiagnosisArena results use 1{,}113 cases, whereas our evaluation uses the 915 publicly available cases.

\paragraph{Supervised EHR baselines.}
We evaluate XGBoost \citep{chen2016xgboost} and TF-IDF with logistic regression as supervised baselines for the six patient-level tasks. For longitudinal MEDS records, XGBoost uses bag-of-codes features, including event counts and last observed values. For emergency department triage, it uses structured triage features and chief-complaint tokens, following established tabular approaches \citep{hong2018triage}. TF-IDF with logistic regression operates directly on the serialized patient state.

Both supervised models are trained using locally labeled examples from the corresponding training partitions. XGBoost tree depth, column subsampling, minimum child weight, and number of trees are selected by five-fold cross-validation. TF-IDF $n$-gram ranges and logistic regression regularization are selected using the same procedure. The selected models are refitted using all available training examples. We evaluate supervised performance with up to 1{,}000 labeled examples per task.

\subsection{Calibration and Evaluation Protocol}
\label{sec:fairness}

We evaluate MedJev in the zero-shot setting without task-specific training or local labels. To examine the calibration of patient-level probabilities, we additionally consider post-hoc recalibration using labeled examples from the corresponding training partitions. For binary \textsc{Noul} tasks, we apply Platt scaling \citep{platt1999probabilistic} to the predicted logits. For multiclass \textsc{Choice} tasks, we apply temperature scaling \citep{guo2017calibration} to the predicted distributions. These transformations use at most two fitted parameters and do not update the underlying language model. For medical knowledge benchmarks, a shared temperature is fitted using MedQA training questions. For patient-level tasks, recalibration is fitted separately for each outcome using local labels.

All models are evaluated on identical test examples. Models requiring labeled data, including supervised baselines and post-hoc recalibration, use the same patient-disjoint training partitions. Model selection and calibration fitting do not use test labels. We additionally vary the available label budget from 25 to 1{,}000 examples to compare the label efficiency of supervised learning and post-hoc recalibration.

For \textsc{Choice} tasks, we report top-1 accuracy and expected calibration error (ECE) computed from the probability assigned to the predicted class using 15 equal-width bins \citep{guo2017calibration}. For binary \textsc{Noul} tasks, we report the area under the receiver operating characteristic curve (AUROC), area under the precision-recall curve (AUPRC), ECE, and Brier score. We compute 95\% percentile bootstrap confidence intervals over test examples \citep{efron1993bootstrap}. Differences between MedJev and directly evaluated baselines are assessed using paired bootstrap resampling of the same test examples.

\paragraph{Implementation.}
All language models are evaluated locally using 8-bit MLX builds \citep{mlx2023} on an Apple silicon laptop with 48\,GB of unified memory. Our inference engine uses the frozen OpenJev helper and replaces only the underlying model client. We verify that local inference reproduces the outputs of the MedJev HTTP server on a verification sample, with a maximum absolute probability difference of zero. MedJev requires approximately 0.5\,s for a MedQA question and 2\,s for a serialized MEDS state. All EHR processing and model inference are performed locally, without transmitting patient records to external services.

\section{Results}
\label{sec:results}

\Cref{tab:mc,tab:binary} summarize the performance of all models across the ten benchmarks. \Cref{tab:delta} reports paired differences relative to zero-shot MedJev. We examine performance on medical knowledge and diagnosis, patient-level prediction, probability calibration, and the effect of local training data.

\begin{table}[t]
  \centering
  \caption{\textbf{Multiple-choice benchmarks:} accuracy (\%) and expected calibration error (ECE) in parentheses. Supervised baselines are evaluated only on ED Triage Disposition. Published results are included for reference. $^\dagger$ indicates open-ended evaluation.}
  \label{tab:mc}
  \small
  \begin{adjustbox}{max width=0.8\linewidth}
    \begin{tabular}{lccccc}
\toprule
Model & MedQA & USMLE & NEJM & DiagnosisArena & ED Triage Disposition \\
\midrule
MedJev (zero-shot) & 72.5 {\scriptsize (0.04)} & 75.4 {\scriptsize (0.03)} & 80.1 {\scriptsize (0.06)} & 55.5 {\scriptsize (0.07)} & 62.6 {\scriptsize (0.14)} \\
\quad + local recalibration & 72.5 {\scriptsize (0.05)} & 75.4 {\scriptsize (0.04)} & 80.1 {\scriptsize (0.06)} & 55.5 {\scriptsize (0.11)} & 62.6 {\scriptsize (0.08)} \\
Qwen3.5-9B (base) & 68.6 {\scriptsize (0.05)} & 70.8 {\scriptsize (0.05)} & 77.0 {\scriptsize (0.04)} & 54.4 {\scriptsize (0.20)} & 62.4 {\scriptsize (0.20)} \\
MedGemma-1.5-4B & 47.7 {\scriptsize (0.30)} & 54.0 {\scriptsize (0.26)} & 60.4 {\scriptsize (0.21)} & 42.3 {\scriptsize (0.38)} & -- \\
Gemma-3-4B & 39.7 {\scriptsize (0.59)} & 47.1 {\scriptsize (0.52)} & 40.6 {\scriptsize (0.58)} & 36.3 {\scriptsize (0.63)} & -- \\
Qwen3.5-2B & 39.4 {\scriptsize (0.16)} & 43.4 {\scriptsize (0.15)} & 48.8 {\scriptsize (0.12)} & 41.7 {\scriptsize (0.24)} & 60.9 {\scriptsize (0.28)} \\
Pythia-2.8B & 19.2 {\scriptsize (0.07)} & 27.4 {\scriptsize (0.01)} & 20.7 {\scriptsize (0.04)} & 23.3 {\scriptsize (0.04)} & 4.2 {\scriptsize (0.14)} \\
XGBoost (supervised) & -- & -- & -- & -- & 70.6 {\scriptsize (0.03)} \\
TF-IDF + LR (supervised) & -- & -- & -- & -- & 67.0 {\scriptsize (0.03)} \\
\midrule
\multicolumn{6}{l}{\textit{Reported in the literature}} \\
GPT-4 & 74.7 & 78.9 & 75.8$^\dagger$ & -- & -- \\
GPT-3.5 & 44.6 & 50.8 & 47.5$^\dagger$ & -- & -- \\
GPT-4o & -- & -- & -- & 36.4 & -- \\
o1 & -- & -- & -- & 61.9 & -- \\
\midrule
Chance & 20.0 & 25.0 & 20.0 & 25.0 & 16.7 \\
\bottomrule
\end{tabular}

  \end{adjustbox}
\end{table}

\begin{table}[t]
  \centering
  \caption{\textbf{Binary EHR benchmarks:} AUROC and ECE in parentheses. Post-hoc recalibration adjusts predicted probabilities without changing AUROC.}
  \label{tab:binary}
  \small
  \begin{adjustbox}{max width=\linewidth}
    \begin{tabular}{lccccc}
\toprule
Model & ED Triage Admission & ED Admission at 2\,h & In-Hospital Mortality & Length of Stay $>$ 7 Days & 30-Day Readmission \\
\midrule
MedJev (zero-shot) & 0.745 {\scriptsize (0.15)} & 0.742 {\scriptsize (0.08)} & 0.926 {\scriptsize (0.29)} & 0.688 {\scriptsize (0.19)} & 0.671 {\scriptsize (0.27)} \\
\quad + local recalibration & 0.745 {\scriptsize (0.04)} & 0.742 {\scriptsize (0.05)} & 0.926 {\scriptsize (0.01)} & 0.688 {\scriptsize (0.03)} & 0.671 {\scriptsize (0.02)} \\
Qwen3.5-9B (base) & 0.756 {\scriptsize (0.22)} & 0.769 {\scriptsize (0.14)} & 0.899 {\scriptsize (0.43)} & 0.664 {\scriptsize (0.35)} & 0.650 {\scriptsize (0.37)} \\
Qwen3.5-2B & 0.694 {\scriptsize (0.29)} & 0.686 {\scriptsize (0.38)} & 0.711 {\scriptsize (0.02)} & 0.421 {\scriptsize (0.15)} & 0.501 {\scriptsize (0.19)} \\
Pythia-2.8B & 0.490 {\scriptsize (0.06)} & 0.631 {\scriptsize (0.07)} & 0.764 {\scriptsize (0.43)} & 0.629 {\scriptsize (0.24)} & 0.555 {\scriptsize (0.28)} \\
XGBoost (supervised) & 0.809 {\scriptsize (0.04)} & 0.824 {\scriptsize (0.04)} & 0.902 {\scriptsize (0.01)} & 0.719 {\scriptsize (0.06)} & 0.684 {\scriptsize (0.05)} \\
TF-IDF + LR (supervised) & 0.742 {\scriptsize (0.05)} & 0.784 {\scriptsize (0.04)} & 0.879 {\scriptsize (0.00)} & 0.734 {\scriptsize (0.04)} & 0.654 {\scriptsize (0.02)} \\
\midrule
Positive rate (\%) & 37.9 & 44.9 & 2.6 & 22.9 & 20.5 \\
\bottomrule
\end{tabular}

  \end{adjustbox}
\end{table}

\subsection{Medical Knowledge and Diagnosis}
\label{sec:res-qa}

\begin{figure}[t]
  \centering
  \includegraphics[width=\linewidth]{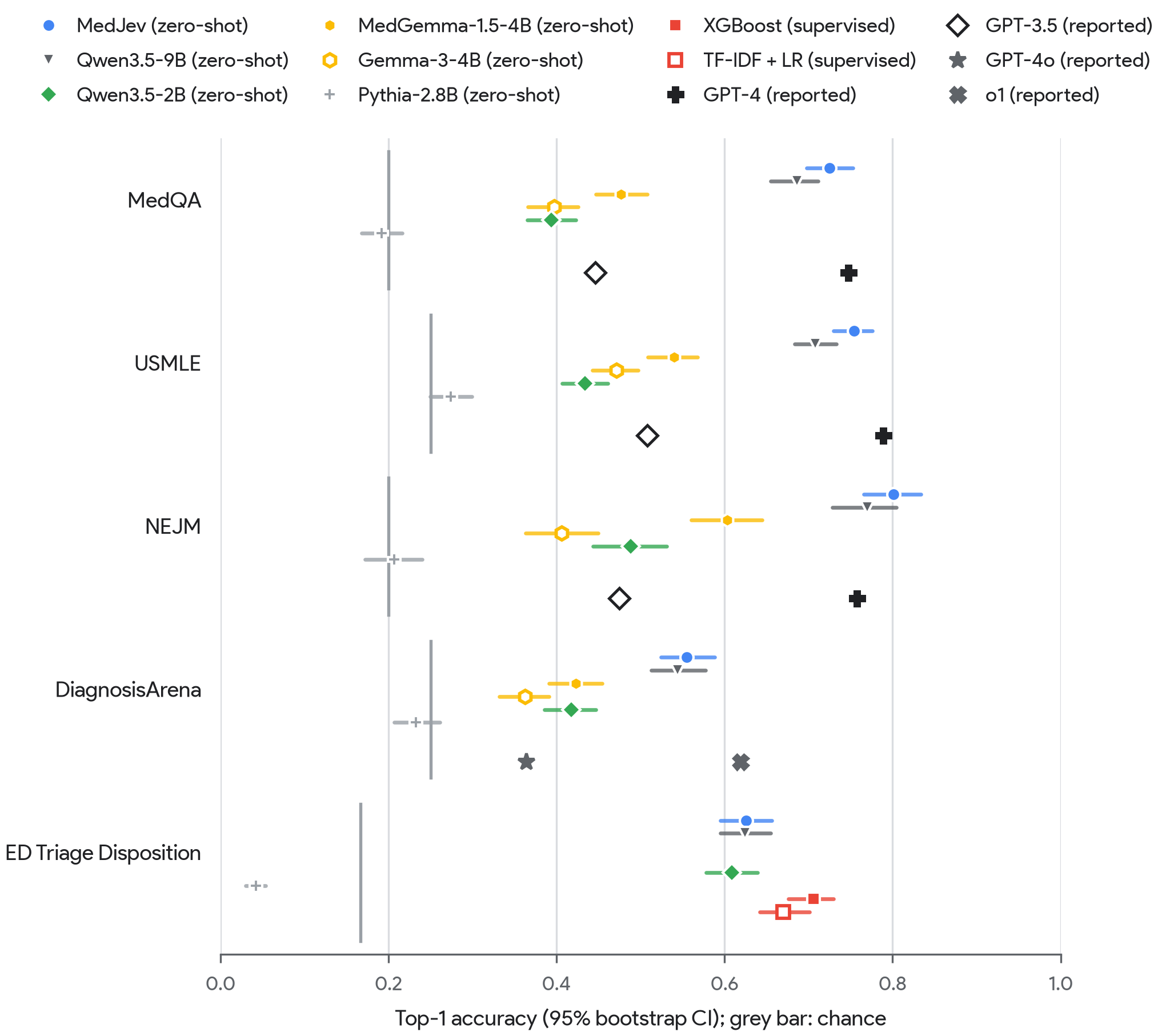}
  \caption{Accuracy on multiple-choice benchmarks with 95\% bootstrap confidence intervals. Grey bars indicate chance performance. Published results include GPT-3.5 and GPT-4 on MedQA and USMLE \citep{nori2023capabilities}, GPT-3.5 and GPT-4 on NEJM \citep{savage2024diagnostic}, and GPT-4o and o1 on DiagnosisArena \citep{zhu2026diagnosisarena}.}
  \label{fig:mc}
\end{figure}

MedJev achieves zero-shot accuracies of \resMedqaJevAcc\% on MedQA (95\% CI \resMedqaJevAccCI), \resUsmleJevAcc\% on USMLE (\resUsmleJevAccCI), \resNejmJevAcc\% on NEJM (\resNejmJevAccCI), and \resDiagarenaJevAcc\% on DiagnosisArena (\resDiagarenaJevAccCI). These results exceed the corresponding chance accuracies of 20--25\% across all four benchmarks (\cref{fig:mc}).

We first compare MedJev with Qwen3.5-9B, its underlying language model. Under identical prompts and probability readouts, Qwen3.5-9B performs \resMedqaBaseDelta{} and \resUsmleBaseDelta{} percentage points below MedJev on MedQA and USMLE, respectively, with paired confidence intervals of \resMedqaBaseDeltaCI{} and \resUsmleBaseDeltaCI. The differences on NEJM (\resNejmBaseDelta) and DiagnosisArena (\resDiagarenaBaseDelta) are smaller and not statistically significant. These comparisons indicate that System-One training improves performance on the medical examination benchmarks, although the improvement is not consistent across all diagnostic tasks.

Model size also affects performance. Qwen3.5-2B achieves accuracies of \resMedqaSmallAcc\%, \resUsmleSmallAcc\%, \resNejmSmallAcc\%, and \resDiagarenaSmallAcc\% on the four benchmarks, respectively. Pythia-2.8B, which does not undergo instruction tuning, achieves \resMedqaPythiaAcc\%, \resUsmlePythiaAcc\%, \resNejmPythiaAcc\%, and \resDiagarenaPythiaAcc\%, with performance near chance across these tasks.

We additionally evaluate MedGemma-1.5-4B and Gemma-3-4B to examine the effect of medical-domain training. MedGemma achieves \resMedqaMedgemmaAcc\% on MedQA, \resUsmleMedgemmaAcc\% on USMLE, \resNejmMedgemmaAcc\% on NEJM, and \resDiagarenaMedgemmaAcc\% on DiagnosisArena. Its accuracy exceeds that of Gemma-3-4B, which achieves \resMedqaGemmaAcc\%, \resUsmleGemmaAcc\%, \resNejmGemmaAcc\%, and \resDiagarenaGemmaAcc\%, respectively. However, MedGemma remains below MedJev on all four benchmarks, with paired differences ranging from \resMedqaMedgemmaDelta{} to \resDiagarenaMedgemmaDelta. MedGemma also exhibits higher calibration error than MedJev, with ECE ranging from \resNejmMedgemmaEce{} to \resDiagarenaMedgemmaEce. Gemma-3-4B has higher ECE, ranging from \resUsmleGemmaEce{} to \resDiagarenaGemmaEce. These results show that medical-domain training improves the performance of Gemma-3-4B, but does not close the gap with MedJev under the probability-based evaluation protocol.

We compare these results with previously reported frontier-model performance. On the five-option MedQA benchmark, GPT-4 achieves \resMedqaRefGptFour\% and GPT-3.5 achieves \resMedqaRefGptThree\% \citep{nori2023capabilities}. On the four-option USMLE benchmark, their reported accuracies are \resUsmleRefGptFour\% and \resUsmleRefGptThree\%, respectively. MedJev performs within 2--4 percentage points of the reported GPT-4 results and substantially exceeds those of GPT-3.5.

On NEJM, MedJev achieves \resNejmJevAcc\% accuracy, compared with reported GPT-4 accuracies of \resNejmGptFourMin--\resNejmGptFourMax\% and GPT-3.5 accuracies of \resNejmGptThreeMin--\resNejmGptThreeMax\% across reasoning prompts \citep{savage2024diagnostic}. However, the published results use open-ended clinician-graded diagnoses, whereas our evaluation uses multiple-choice questions. These results are therefore not directly comparable. On DiagnosisArena, MedJev achieves \resDiagarenaJevAcc\%, compared with reported accuracies of \resDiagarenaRefGptFourO\% for GPT-4o and \resDiagarenaRefOone\% for o1 \citep{zhu2026diagnosisarena}. The published results use the full benchmark, while our evaluation uses its 915 publicly available cases.

\subsection{Patient-Level Prediction from EHRs}
\label{sec:res-ehr}

\begin{figure}[t]
  \centering
  \includegraphics[width=\linewidth]{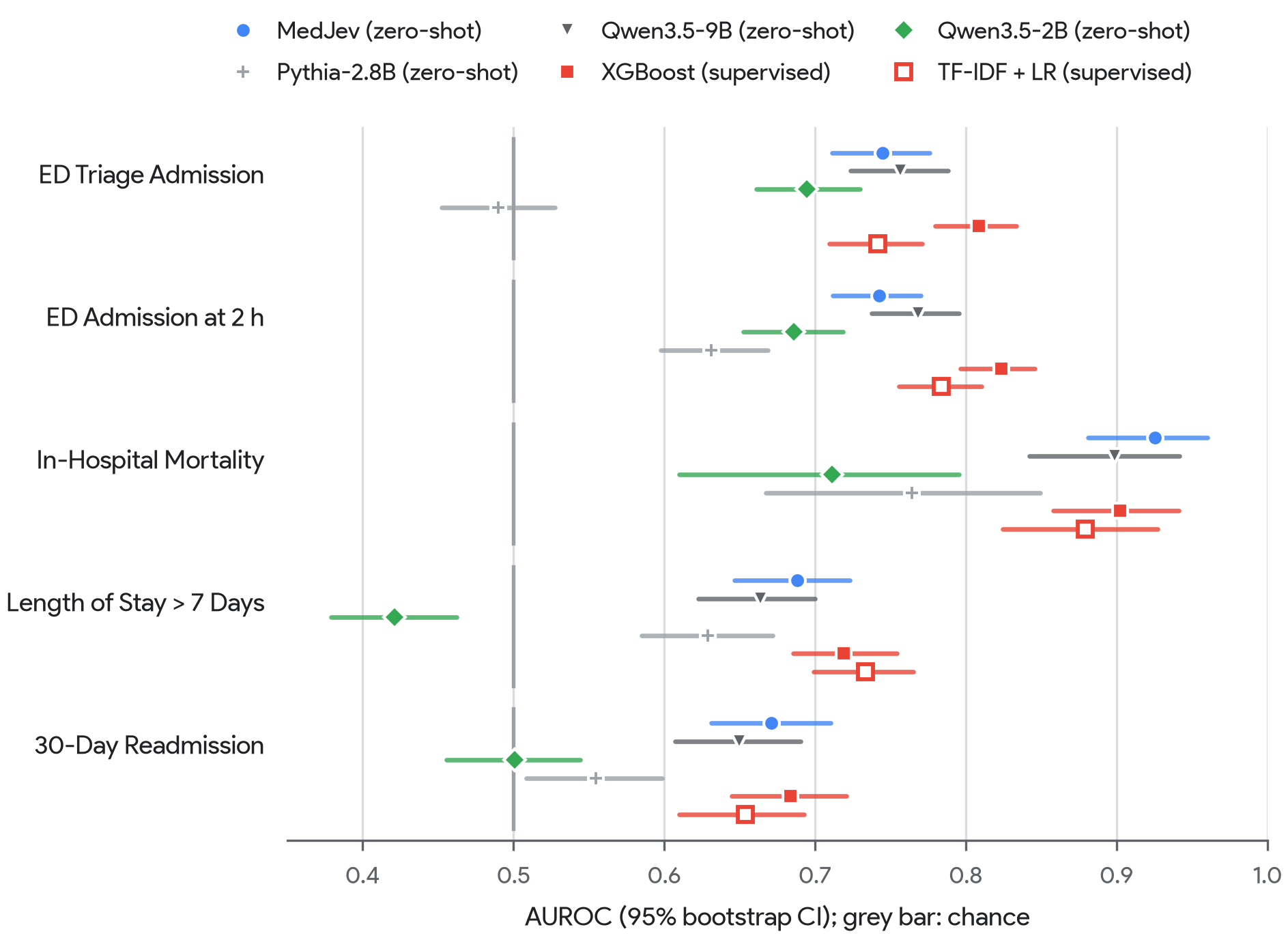}
  \caption{\textbf{AUROC on binary EHR benchmarks with 95\% bootstrap confidence intervals}. Zero-shot models are evaluated without task-specific labels. Supervised baselines are trained using local labeled examples. All models are evaluated on the same test examples.}
  \label{fig:auroc}
\end{figure}

\begin{table}[t]
  \centering
  \caption{\textbf{Paired performance differences relative to zero-shot MedJev with 95\% bootstrap confidence intervals.} Differences are measured in AUROC for binary tasks and percentage points for multiple-choice tasks. Negative values indicate lower performance than MedJev.}
  \label{tab:delta}
  \small
  \resizebox{\linewidth}{!}{\begin{tabular}{lccccccc}
\toprule
Benchmark & Qwen3.5-9B (base) & MedGemma-1.5-4B & Gemma-3-4B & Qwen3.5-2B & Pythia-2.8B & XGBoost & TF-IDF + LR \\
\midrule
MedQA (accuracy pts) & -3.9 {\scriptsize [-6.3 to -1.4]} & -24.8 {\scriptsize [-28.4 to -20.9]} & -32.8 {\scriptsize [-36.4 to -28.8]} & -33.1 {\scriptsize [-36.6 to -29.6]} & -53.3 {\scriptsize [-57.0 to -49.3]} & -- & -- \\
USMLE (accuracy pts) & -4.6 {\scriptsize [-6.8 to -2.4]} & -21.4 {\scriptsize [-24.4 to -18.5]} & -28.3 {\scriptsize [-31.6 to -24.6]} & -32.1 {\scriptsize [-35.2 to -28.8]} & -48.0 {\scriptsize [-51.1 to -44.7]} & -- & -- \\
NEJM (accuracy pts) & -3.1 {\scriptsize [-6.6 to +0.2]} & -19.7 {\scriptsize [-24.4 to -15.2]} & -39.5 {\scriptsize [-44.7 to -34.4]} & -31.2 {\scriptsize [-36.1 to -26.6]} & -59.4 {\scriptsize [-64.3 to -54.1]} & -- & -- \\
DiagnosisArena (accuracy pts) & -1.1 {\scriptsize [-3.9 to +1.9]} & -13.2 {\scriptsize [-17.3 to -9.3]} & -19.2 {\scriptsize [-23.0 to -15.2]} & -13.8 {\scriptsize [-17.3 to -10.2]} & -32.2 {\scriptsize [-36.4 to -28.2]} & -- & -- \\
ED Triage Disposition (accuracy pts) & -0.2 {\scriptsize [-1.9 to +1.5]} & -- & -- & -1.7 {\scriptsize [-4.3 to +1.0]} & -58.4 {\scriptsize [-61.6 to -54.9]} & +8.0 {\scriptsize [+4.9 to +11.1]} & +4.4 {\scriptsize [+1.1 to +7.7]} \\
ED Triage Admission (AUROC) & +0.012 {\scriptsize [-0.003 to +0.027]} & -- & -- & -0.050 {\scriptsize [-0.077 to -0.024]} & -0.255 {\scriptsize [-0.303 to -0.206]} & +0.064 {\scriptsize [+0.034 to +0.093]} & -0.003 {\scriptsize [-0.039 to +0.031]} \\
ED Admission at 2\,h (AUROC) & +0.026 {\scriptsize [+0.007 to +0.045]} & -- & -- & -0.056 {\scriptsize [-0.089 to -0.025]} & -0.111 {\scriptsize [-0.149 to -0.074]} & +0.081 {\scriptsize [+0.053 to +0.113]} & +0.042 {\scriptsize [+0.010 to +0.074]} \\
In-Hospital Mortality (AUROC) & -0.027 {\scriptsize [-0.054 to -0.006]} & -- & -- & -0.214 {\scriptsize [-0.294 to -0.136]} & -0.161 {\scriptsize [-0.258 to -0.064]} & -0.023 {\scriptsize [-0.054 to +0.006]} & -0.046 {\scriptsize [-0.089 to -0.007]} \\
Length of Stay $>$ 7 Days (AUROC) & -0.024 {\scriptsize [-0.045 to -0.006]} & -- & -- & -0.267 {\scriptsize [-0.323 to -0.206]} & -0.059 {\scriptsize [-0.104 to -0.019]} & +0.031 {\scriptsize [-0.009 to +0.069]} & +0.045 {\scriptsize [+0.011 to +0.080]} \\
30-Day Readmission (AUROC) & -0.022 {\scriptsize [-0.047 to +0.004]} & -- & -- & -0.171 {\scriptsize [-0.225 to -0.111]} & -0.117 {\scriptsize [-0.165 to -0.063]} & +0.012 {\scriptsize [-0.028 to +0.053]} & -0.017 {\scriptsize [-0.052 to +0.018]} \\
\bottomrule
\end{tabular}
}
\end{table}

\Cref{fig:auroc,tab:delta} compare MedJev with zero-shot language models and supervised baselines on patient-level prediction tasks. Performance varies by outcome, with stronger zero-shot results for clinical severity prediction than for hospital operational outcomes.

On In-Hospital Mortality, MedJev achieves an AUROC of \resMortalityJevAuroc{} (95\% CI \resMortalityJevAurocCI). XGBoost trained on labeled admissions achieves \resMortalityXgbAuroc{}, with a paired difference of \resMortalityXgbDelta{} (95\% CI \resMortalityXgbDeltaCI). TF-IDF with logistic regression achieves \resMortalityTfidfAuroc{}, with a paired difference of \resMortalityTfidfDelta{} (95\% CI \resMortalityTfidfDeltaCI). MedJev therefore achieves mortality discrimination comparable to or better than the supervised baselines without task-specific training. On 30-Day Readmission, MedJev and the supervised baselines achieve AUROCs between 0.65 and 0.69. The difference between XGBoost and MedJev is \resReadmissionXgbDelta{} (95\% CI \resReadmissionXgbDeltaCI), indicating no statistically significant improvement from supervision at this label budget.

Supervised models perform better on several operational outcomes. On ED Triage Admission, XGBoost exceeds MedJev by \resTriageXgbDelta{} AUROC (95\% CI \resTriageXgbDeltaCI). The corresponding difference for ED Admission at 2\,h is \resEdmedsXgbDelta{} (95\% CI \resEdmedsXgbDeltaCI). On ED Triage Disposition, XGBoost achieves an accuracy improvement of \resTriageDispXgbDelta{} percentage points (95\% CI \resTriageDispXgbDeltaCI). For Length of Stay $>$ 7 Days, TF-IDF exceeds MedJev by \resLongLOSTfidfDelta{} AUROC (95\% CI \resLongLOSTfidfDeltaCI). These results indicate that supervised models are more effective for outcomes associated with local hospital operations.

Among zero-shot models, MedJev outperforms its base model on In-Hospital Mortality and Length of Stay $>$ 7 Days, with paired differences of \resMortalityBaseDelta{} and \resLongLOSBaseDelta{}, respectively. However, Qwen3.5-9B performs better on ED Admission at 2\,h, with a difference of \resEdmedsBaseDelta{}. Qwen3.5-2B achieves an AUROC of \resMortalitySmallAuroc{} on In-Hospital Mortality, \resReadmissionSmallAuroc{} on 30-Day Readmission, and \resLongLOSSmallAuroc{} on Length of Stay $>$ 7 Days. Pythia-2.8B achieves AUROCs of \resMortalityPythiaAuroc{} and \resLongLOSPythiaAuroc{} on In-Hospital Mortality and Length of Stay $>$ 7 Days, respectively, but performs near chance on ED Triage Admission (\resTriagePythiaAuroc). Its accuracy on ED Triage Disposition is \resTriageDispPythiaAcc\%. Overall, MedJev achieves stronger performance than the smaller zero-shot models on most patient-level tasks, although System-One training does not consistently improve performance over its base model.

\subsection{Probability Calibration}
\label{sec:res-cal}

\begin{figure}[t]
  \centering
  \includegraphics[width=\linewidth]{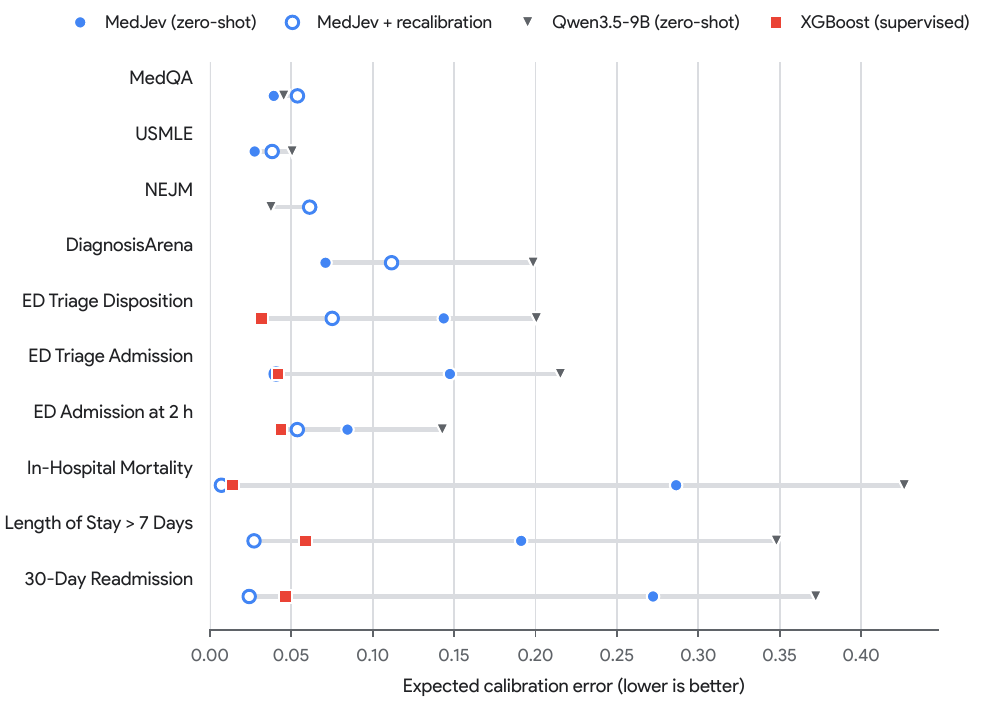}
  \caption{\textbf{Expected calibration error across the ten benchmarks for zero-shot MedJev, recalibrated MedJev, Qwen3.5-9B, and XGBoost. Lower values indicate better calibration.}}
  \label{fig:ece}
\end{figure}

\begin{figure}[t]
  \centering
  \includegraphics[width=\linewidth]{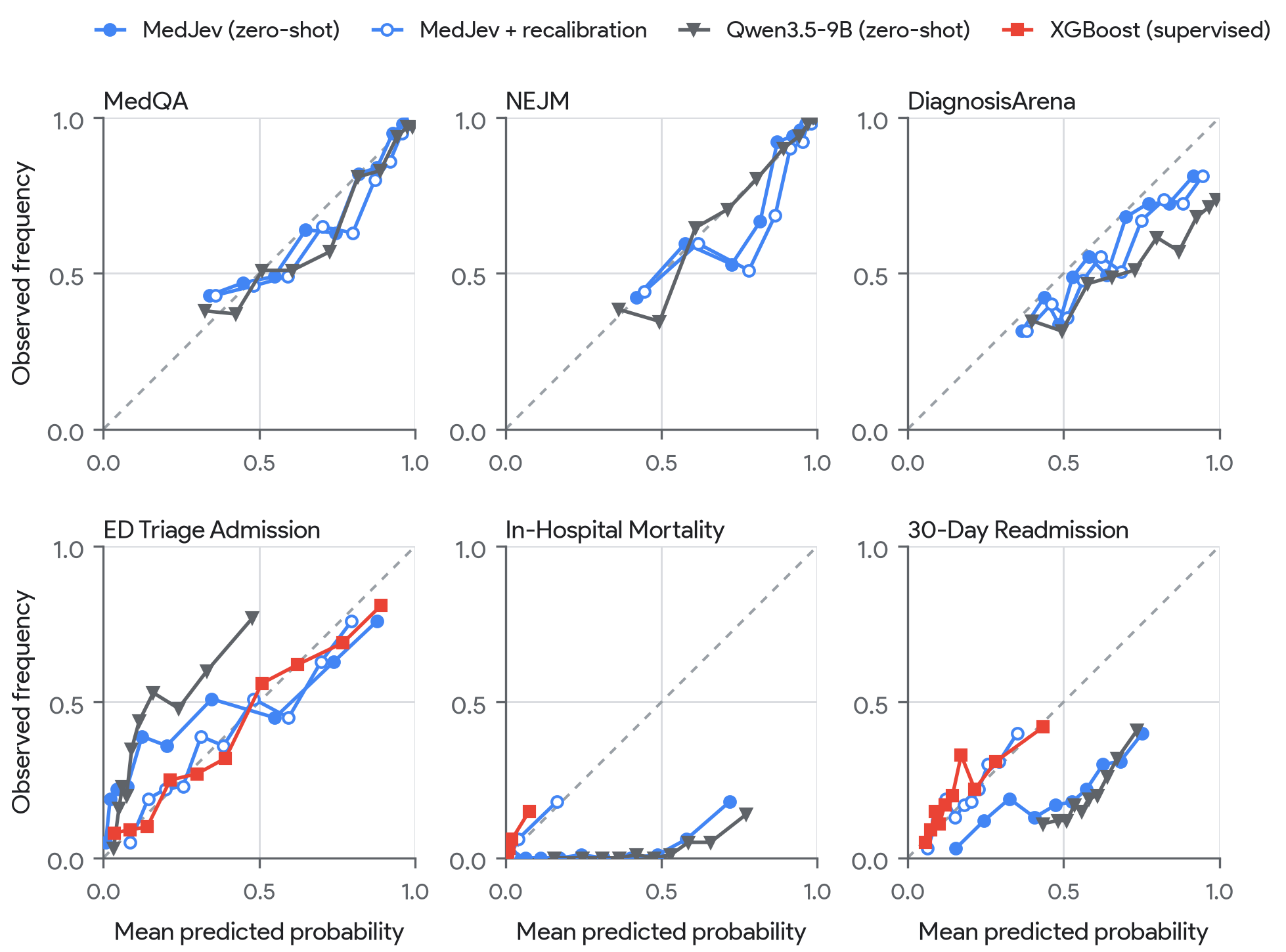}
  \caption{\textbf{Calibration curves based on prediction deciles.} For multiple-choice tasks, observed accuracy is plotted against predicted confidence. For binary tasks, observed outcome frequency is plotted against predicted probability. The dashed diagonal indicates perfect calibration.}
  \label{fig:calcurves}
\end{figure}

We evaluate whether the probabilities produced by MedJev correspond to observed outcome frequencies. \Cref{fig:ece,fig:calcurves} present expected calibration error and calibration curves across the benchmarks.

MedJev is well calibrated on medical knowledge benchmarks, with ECE values of \resMedqaJevEce{} on MedQA, \resUsmleJevEce{} on USMLE, and \resNejmJevEce{} on NEJM. In contrast, its zero-shot probabilities are poorly calibrated on patient-level outcomes. For In-Hospital Mortality, the mean predicted probability is \resMortalityJevMeanProb{}, compared with an observed mortality rate of \resMortalityPrev\%, resulting in an ECE of \resMortalityJevEce{}. The corresponding ECE values are \resLongLOSJevEce{} for Length of Stay $>$ 7 Days, \resReadmissionJevEce{} for 30-Day Readmission, and \resTriageJevEce{} for ED Triage Admission. These results indicate that good risk discrimination does not necessarily imply calibrated outcome probabilities in the zero-shot setting.

We apply post-hoc recalibration using labeled examples from the corresponding training partitions. Recalibration reduces ECE to \resMortalityPlattEce{} on In-Hospital Mortality, \resLongLOSPlattEce{} on Length of Stay $>$ 7 Days, \resReadmissionPlattEce{} on 30-Day Readmission, and \resTriagePlattEce{} on ED Triage Admission. These values are comparable to those obtained by XGBoost, which achieves ECE values of \resMortalityXgbEce{}, \resLongLOSXgbEce{}, \resReadmissionXgbEce{}, and \resTriageXgbEce{} on the same tasks. Recalibration changes the probability estimates without changing the reported AUROC.

We also examine the effect of System-One training on calibration. Qwen3.5-9B has ECE values of \resMortalityBaseEce{}, \resLongLOSBaseEce{}, and \resReadmissionBaseEce{} on In-Hospital Mortality, Length of Stay $>$ 7 Days, and 30-Day Readmission, respectively. These values are higher than those of MedJev, indicating that System-One training improves probability calibration even when local recalibration remains necessary.

Calibration parameters do not consistently transfer across benchmarks. Applying a temperature fitted on MedQA training questions changes ECE from \resUsmleJevEce{} to \resUsmlePlattEce{} on USMLE and from \resNejmJevEce{} to \resNejmPlattEce{} on NEJM. On DiagnosisArena, ECE increases from \resDiagarenaJevEce{} to \resDiagarenaPlattEce{}. These results indicate that calibration fitted on one benchmark does not necessarily improve probability estimates on another.

\subsection{Effect of Local Training Data}
\label{sec:res-le}

\begin{figure}[t]
  \centering
  \includegraphics[width=\linewidth]{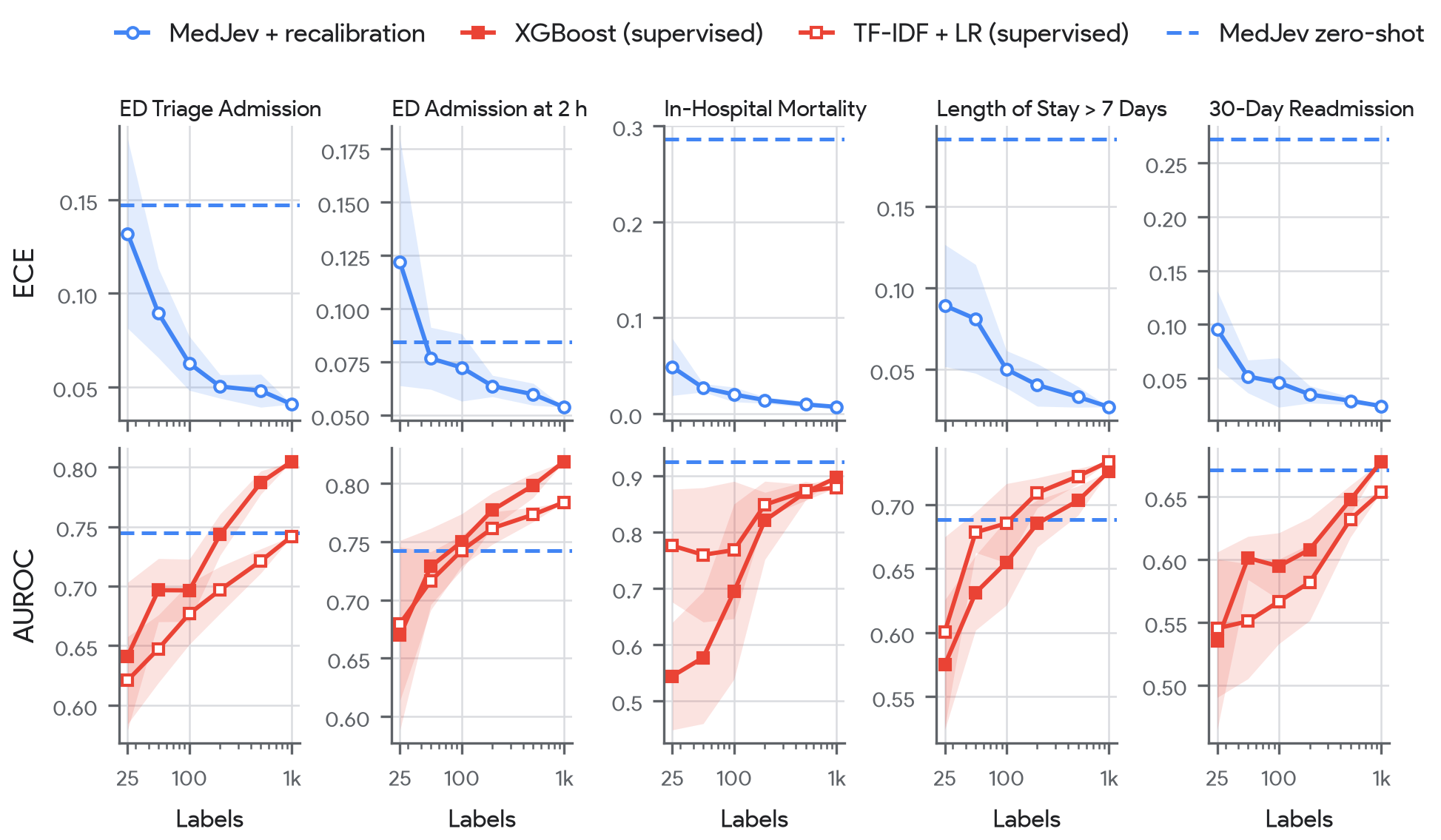}
  \caption{\textbf{Effect of the number of local labels on probability calibration and supervised prediction performance.} The upper panels show ECE for recalibrated MedJev, and the lower panels show AUROC for XGBoost and TF-IDF with logistic regression. Dashed lines indicate zero-shot MedJev performance. Results are reported as mean $\pm$ standard deviation over 10 random draws per label budget.}
  \label{fig:le}
\end{figure}

We vary the number of labeled training examples from 25 to 1{,}000 to assess the data requirements of recalibration and supervised learning (\cref{fig:le}). For MedJev, we fit post-hoc calibration parameters using each label budget while keeping the model predictions fixed. For the supervised baselines, we train separate models using the corresponding labeled examples.

Recalibration improves probability estimates with relatively few labeled examples. On In-Hospital Mortality, Platt scaling reduces ECE from \resMortalityJevEce{} to approximately 0.03 using 50 labeled examples. With 200 examples, ECE ranges from 0.014 to 0.064 across the binary patient-level benchmarks. Because recalibration does not change the reported AUROC, these improvements are obtained without modifying the underlying model or its discrimination performance.

The number of labels required for supervised models to match zero-shot MedJev varies across tasks. XGBoost reaches comparable AUROC with approximately 100 labels on ED Admission at 2\,h and 200 labels on ED Triage Admission. TF-IDF requires approximately 200 labels to match MedJev on Length of Stay $>$ 7 Days. On 30-Day Readmission, comparable performance is reached with approximately 1{,}000 labels. For In-Hospital Mortality, neither supervised baseline matches the zero-shot AUROC of MedJev within the evaluated label budgets.

These results indicate that the relative benefit of zero-shot prediction depends on the task and the availability of local labels. Recalibration requires fewer labels than supervised training to improve probability estimates, while supervised models can provide better discrimination for some outcomes when sufficient labeled examples are available.

\subsection{Selective Prediction}
\label{sec:res-sel}

\begin{figure}[t]
  \centering
  \includegraphics[width=\linewidth]{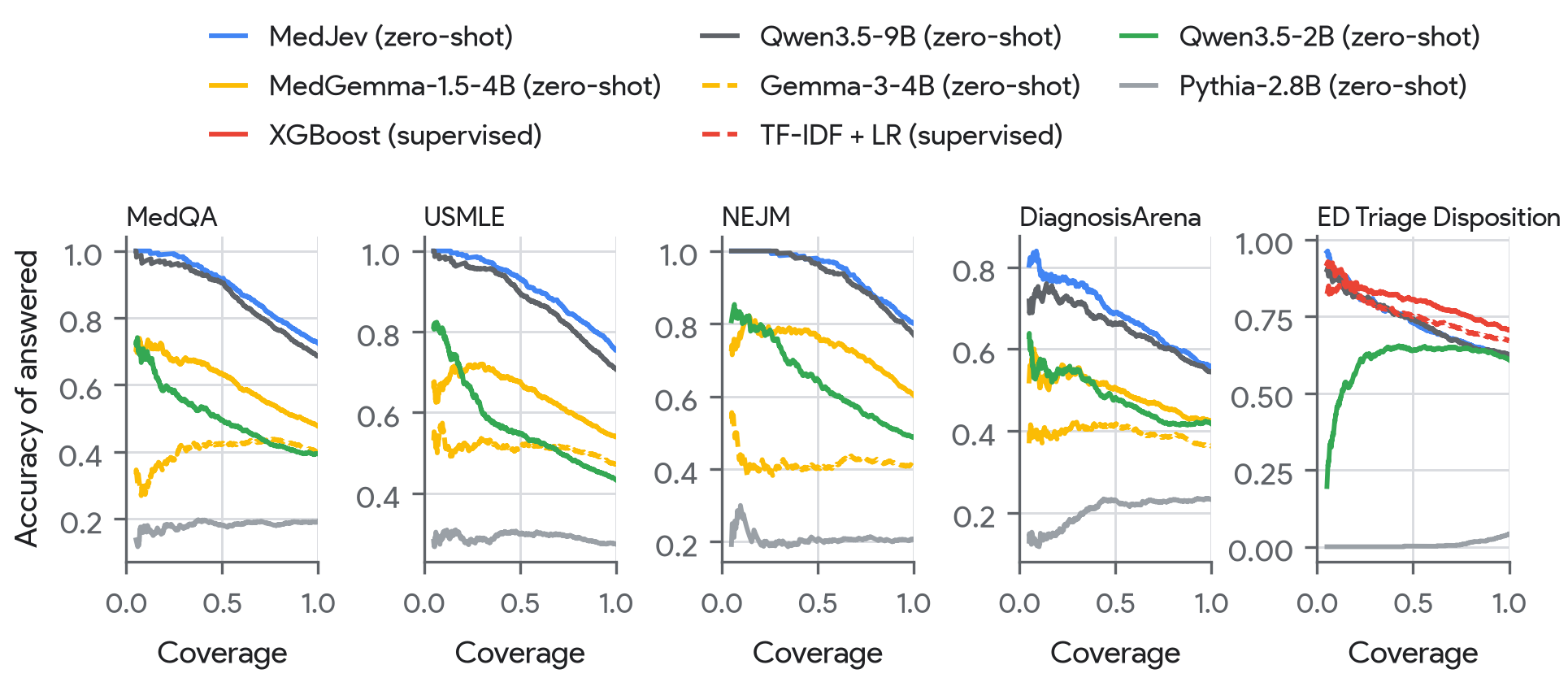}
  \caption{\textbf{Accuracy as a function of coverage on the multiple-choice benchmarks.} Coverage denotes the fraction of examples retained after selecting predictions with the highest confidence.}
  \label{fig:selective}
\end{figure}

We evaluate selective prediction by ranking examples according to the probability assigned to the predicted class and measuring accuracy at different coverage levels. This analysis examines whether model confidence can identify predictions that are more likely to be correct.

As shown in \cref{fig:selective}, MedJev generally achieves higher accuracy as coverage decreases on the medical knowledge and diagnosis benchmarks. Its accuracy also exceeds that of Qwen3.5-9B across most coverage levels. These results indicate that MedJev's confidence estimates contain information about prediction correctness and can support confidence-based selection of model outputs.

\section{Discussion}
\label{sec:discussion}

Our results show that System-One models can perform medical decision tasks without  autoregressive generation. MedJev achieves competitive performance on medical knowledge and diagnosis benchmarks and provides useful discrimination for several patient-level outcomes. However, its performance varies across clinical tasks, and its zero-shot probability estimates do not consistently reflect local outcome frequencies. These findings suggest that System-One models can serve as zero-shot baselines for medical prediction, with local recalibration or supervised training required depending on the task.

\paragraph{Zero-shot predictive performance.}
MedJev achieves strong accuracy and calibration on medical knowledge benchmarks using a single forward pass. Its performance on MedQA and USMLE is within a few percentage points of published zero-shot GPT-4 results and substantially exceeds reported GPT-3.5 accuracy. On DiagnosisArena, MedJev outperforms the reported GPT-4o result but remains below o1. These comparisons are limited by differences in evaluation protocols, particularly for benchmarks originally evaluated using open-ended responses. Nevertheless, the results demonstrate that structured probability estimation can provide competitive performance on medical question-answering tasks without generating textual answers.

On patient-level prediction, MedJev achieves an AUROC of \resMortalityJevAuroc{} for In-Hospital Mortality, comparable to supervised models trained on labeled examples. Performance is also comparable to supervised baselines on 30-Day Readmission. In contrast, supervised models perform better on ED admission and length-of-stay prediction. These differences may reflect the extent to which outcomes depend on patient characteristics and local clinical practices. Mortality is closely associated with clinical severity, whereas admission and discharge decisions are also influenced by institutional policies, resource availability, and care processes. Consequently, the relative performance of zero-shot and supervised models depends on the prediction target.

Comparisons with Qwen3.5-9B further indicate that System-One training contributes to predictive performance. MedJev improves accuracy on MedQA and USMLE and discrimination on In-Hospital Mortality and Length of Stay $>$ 7 Days under the same prompting and readout procedure. It also produces better-calibrated probabilities than its base model across several benchmarks. However, these improvements are not consistent across all tasks, as Qwen3.5-9B performs better on ED admission prediction. System-One training therefore improves performance on selected medical tasks but does not uniformly outperform the underlying language model.

\paragraph{Probability calibration and local adaptation.}
Although MedJev produces well-calibrated confidence estimates on medical knowledge benchmarks, its patient-level predictions exhibit substantial calibration error. In particular, the model overestimates the probability of in-hospital mortality relative to the observed event rate. This distinction is important because discrimination measures whether a model correctly ranks patients by risk, whereas calibration measures whether predicted probabilities correspond to observed outcome frequencies. A model can achieve high AUROC while producing probabilities that are unsuitable for direct interpretation as patient risk.

Post-hoc recalibration substantially reduces this error without retraining MedJev. Using Platt scaling fitted on local labeled examples, we obtain calibration performance comparable to supervised baselines while preserving the original risk ranking. The label-efficiency experiments show that 50--200 labeled examples can substantially improve calibration across patient-level tasks. In comparison, supervised models generally require more labeled examples to match the discrimination of zero-shot MedJev, although this varies by outcome. Calibration parameters fitted on one medical knowledge benchmark do not consistently improve calibration on other benchmarks, indicating that calibration depends on the target task and population.

These results support separating zero-shot prediction from local probability calibration in the evaluation pipeline. The System-One model provides an initial prediction without task-specific training, while a small labeled dataset can be used to adjust the probability estimates to the local population. However, recalibration does not improve discrimination and cannot address limitations in the underlying risk ranking. For tasks where supervised models achieve higher AUROC, additional local training remains beneficial.

\paragraph{Implications for medical evaluation.}
The structured request interface allows medical knowledge questions and patient-level prediction tasks to be evaluated using the same inference procedure. Each task is represented by an input state, a question, and a set of possible outcomes, and the model returns probabilities that can be evaluated for accuracy, discrimination, and calibration. This removes the need to generate and parse textual responses and allows direct comparisons with other language models using the same probability readout.

The results also indicate that evaluation of System-One models should consider both zero-shot performance and the effect of local adaptation. Zero-shot evaluation measures predictive performance without task-specific labels, while recalibration measures whether the predicted probabilities can be adjusted using limited local data. Supervised baselines provide a reference for the performance achievable through task-specific training. Considering these settings together helps determine when a System-One model provides a useful alternative to conventional predictive models. The local inference setup further allows these evaluations to be conducted without transmitting patient records to external model services.

\paragraph{Limitations.}
Several limitations should be considered when interpreting these results. First, the number of positive outcomes is small for some tasks. In-Hospital Mortality contains only 2.6\% deaths in the test set, resulting in wide confidence intervals for AUROC. Second, we evaluate a single prompt and serialization format for each task. The questions, input representations, and event limits are fixed rather than optimized, although prior work has shown that serialization choices can affect language-model performance on structured data \citep{hegselmann2023tabllm}. Third, the NEJM benchmark was originally evaluated using open-ended diagnoses, whereas our evaluation uses multiple-choice questions. Published GPT results therefore provide contextual references rather than directly comparable estimates.

The medical knowledge benchmarks also have potential overlap with model pretraining data. MedQA and USMLE share question stems, and MedQA has been publicly available since 2020. We cannot exclude the possibility that some evaluation questions appeared during pretraining. For the EHR benchmarks, the MEDS extract does not provide a complete ICU item dictionary. We identify bedside measurements using standard MIMIC-IV item identifiers, but omit other chart events that cannot be reliably interpreted. Finally, MIMIC-IV and MIMIC-IV-ED originate from a single academic medical center. The results may not generalize to institutions with different patient populations, clinical practices, or outcome prevalence. Evaluation on additional clinical datasets is needed to assess generalizability.

\paragraph{Data use and ethics.}
MIMIC-IV and MIMIC-IV-ED were accessed under the PhysioNet credentialed data use agreement. All data processing and model inference were performed locally, with the model server restricted to the loopback interface. No individual-level patient records are included in this paper, and calibration parameters fitted on credentialed data are not publicly released. The evaluated models are research artifacts and have not been validated for clinical deployment. Their predictions should not be used to guide patient care.

\bibliographystyle{unsrt}
\bibliography{main}

 \appendix
\setcounter{figure}{0}
\renewcommand{\figurename}{Supplementary Figure}
\renewcommand{\thefigure}{\arabic{figure}}
\renewcommand{\theHfigure}{suppfigure.\arabic{figure}}

\setcounter{table}{0}
\renewcommand{\tablename}{Supplementary Table}
\renewcommand{\thetable}{\arabic{table}}
\renewcommand{\theHtable}{supptable.\arabic{table}}

\clearpage

\startcontents[appendices]

\begin{center}
  \Large \textbf{Appendix Table of Contents}
\end{center}

\printcontents[appendices]{l}{1}{\setcounter{tocdepth}{3}}

\newpage
\appendix
\section*{Appendix}

\section{Question Formats}
\label{app:prompts}

All benchmarks are represented as Jev requests consisting of a state and a typed question. \textsc{Choice} questions return a probability distribution over candidate answers, while \textsc{Noul} questions return the probability of a binary outcome. Table~\ref{tab:app-questions} summarizes the questions used for each benchmark.

\begin{table}[h!]
\centering
\small
\renewcommand{\arraystretch}{1.35}
\begin{tabular}{p{0.23\linewidth}p{0.12\linewidth}p{0.57\linewidth}}
\toprule
Benchmark & Type & Question or outcome \\
\midrule
MedQA, USMLE, NEJM
& \textsc{Choice}
& Which option correctly answers the question? Five options for MedQA and NEJM, four for USMLE. \\

DiagnosisArena
& \textsc{Choice}
& Which option is the most likely final diagnosis for the patient described in the state? Four candidate diagnoses. \\

\midrule
In-Hospital Mortality
& \textsc{Noul}
& Will this patient die before being discharged from the current hospital admission? \\

Length of Stay $>$ 7 Days
& \textsc{Noul}
& Will the current hospital stay last longer than 7 days in total? \\

30-Day Readmission
& \textsc{Noul}
& Will the patient be admitted again within 30 days after this discharge? \\

ED Admission at 2\,h
& \textsc{Noul}
& Based on the information available 2 hours after arrival, will this ED visit end in admission to the hospital? \\

ED Triage Admission
& \textsc{Noul}
& Based on the information available at triage, will this visit end with the patient admitted to the hospital? \\

ED Triage Disposition
& \textsc{Choice}
& What will the disposition of this ED visit be? Options: home, admitted, transfer, left before completion, expired, or other. \\
\bottomrule
\end{tabular}
\caption{Question formats used across the ten medical benchmarks.}
\label{tab:app-questions}
\end{table}

\paragraph{Prompt construction.}
Each question is rendered as a prompt containing the serialized state, question instructions, and candidate options. For \textsc{Choice} questions, the options are assigned single-token letter labels. The prompt follows the format below.

\begin{verbatim}
State:
<state JSON>

Question:
<instructions>

Options:
[A] <option 1>: <description>
[B] <option 2>: <description>
...

Answer with the letter of the best option only.
\end{verbatim}

For \textsc{Noul} questions, the candidate outcomes correspond to true and false, with optional descriptions specifying each outcome. MedJev reads the option-token probabilities at the first answer position and applies the transformations in \cref{eq:readout,eq:noul} to obtain the final probability distribution.

\section{Worked examples}
\label{app:examples}

One example per benchmark, with MedJev's actual zero-shot output. For the knowledge benchmarks we show a real test
item: the one at the 10th percentile of question length, so it is neither hand-picked nor trivially short, together
with its gold answer and MedJev's full option distribution (long DiagnosisArena fields are abridged with
\texttt{[...]}). Individual-level MIMIC data may not be published, so the EHR examples use \emph{synthetic} patients
written in exactly each benchmark's state format; MedJev answered them live, and they have no gold outcome. For a
\textsc{Noul}, \texttt{p(yes)} is the probability that the statement is true.

\par\medskip\noindent\begin{minipage}{\linewidth}\textbf{MedQA}\par\smallskip
\begin{lstlisting}
[MedQA]  real test item test/420
STATE
{
 "subject": "medicine (USMLE)",
 "question": "A 15-year-old boy is brought to the physician for evaluation of a learning disability. His
     teachers have reported hyperactivity during class, difficulties with social interaction, and poor scores on
     reading and writing assessments. Molecular analysis shows an increased number of CGG trinucleotide repeats.
     Which of the following findings are most likely to be seen on physical examination of this patient?"
}
QUESTION (Choice): Which option correctly answers `question`?
  A  Frontal balding and cataracts
  B  Long face and large everted ears
  C  Almond-shaped eyes and downturned mouth
  D  Flat facies and protruding tongue
  E  Thin upper lip and receding chin
GOLD: B
MEDJEV: B  (p = 0.41)  -> correct
  probabilities: A 0.08  B 0.41  C 0.12  D 0.26  E 0.13
\end{lstlisting}
\end{minipage}

\par\medskip\noindent\begin{minipage}{\linewidth}\textbf{USMLE}\par\smallskip
\begin{lstlisting}
[USMLE]  real test item 296
STATE
{
 "subject": "medicine (USMLE)",
 "question": "A 52-year-old fisherman presents to the clinic for an evaluation of a pigmented lesion on his
     neck. He states that he first noticed the lesion last year, but he believes that it has been slowly growing
     in size. Dermatopathology determines that the lesion contains neoplastic cells of melanocytic origin. Which
     of the following characteristics of the lesion would likely be found on physical examination?"
}
QUESTION (Choice): Which option correctly answers `question`?
  A  Brown-black color in one area of the lesion to red-white in a different area
  B  Macule that is 5mm in diameter
  C  Well-circumscribed papule with smooth borders
  D  Itching and pain to palpation
GOLD: A
MEDJEV: A  (p = 0.87)  -> correct
  probabilities: A 0.87  B 0.06  C 0.03  D 0.04
\end{lstlisting}
\end{minipage}

\par\medskip\noindent\begin{minipage}{\linewidth}\textbf{NEJM}\par\smallskip
\begin{lstlisting}
[NEJM]  real test item 202
STATE
{
 "question": "A 32-year-old G1P0 woman presents to her obstetrician for a prenatal visit. She is 30 weeks
     pregnant. She reports some fatigue and complains of urinary urgency. Prior to this pregnancy, she had no
     significant medical history. She takes a prenatal vitamin and folate supplements daily. Her mother has
     diabetes, and her brother has coronary artery disease. On physical examination, the fundal height is 25
     centimeters. A fetal ultrasound shows a proportional reduction in head circumference, trunk size, and limb
     length. What is the most likely cause of the patient's presentation?"
}
QUESTION (Choice): Which option correctly answers `question`?
  A  Antiphospholipid syndrome
  B  Cigarette smoking
  C  Gestational diabetes
  D  Pre-eclampsia
  E  Rubella infection
GOLD: E
MEDJEV: B  (p = 0.85)  -> wrong
  probabilities: A 0.04  B 0.85  C 0.06  D 0.02  E 0.04
\end{lstlisting}
\end{minipage}

\par\medskip\noindent\begin{minipage}{\linewidth}\textbf{DiagnosisArena}\par\smallskip
\begin{lstlisting}
[DiagnosisArena]  real test item 150
STATE
{
 "case_information": "A woman in her 50s presented with bilateral nasal obstruction, thick nasal discharge, and
     facial pain and pressure. Her condition was recalcitrant to 3 prior surgeries, immunotherapy, and medical
     therapy.",
 "physical_examination": "Initial post-procedure examination showed:- Binocular horizontal diplopia with left
     lateral gaze- Inability for the left eye to track beyond the midline- Left eye was not proptotic,
     ecchymotic, or tense- Bedside intraocular pressure measurements: 15 OD, 16 OS (normal)- National Institutes
     of Health stroke score of 1",
 "diagnostic_tests": "{'Clinical Images': {'Figure 1': {'Title': 'Clinical Images', 'Description': 'Images
     showing left-sided abducens nerve palsy immediately following surgery and its resolution by postoperative
     day 1'}, 'Figure 2': {'Title': 'Illustrative Representations of the Proposed Mechanisms', 'Description':
     'Diagrams showing different hypotheses for abducens ne [...]"
}
QUESTION (Choice): Which option is the most likely final diagnosis for the patient described in the state?
  A  Allergic fungal rhinosinusitis
  B  Transient left abducens (cranial nerve VI) nerve palsy
  C  Chronic rhinosinusitis with nasal polyps
  D  Iatrogenic Abducens Nerve Palsy
GOLD: B
MEDJEV: B  (p = 0.70)  -> correct
  probabilities: A 0.01  B 0.70  C 0.01  D 0.28
\end{lstlisting}
\end{minipage}

\par\medskip\noindent\begin{minipage}{\linewidth}\textbf{ED Triage Admission}\par\smallskip
\begin{lstlisting}
[ED Triage Admission]  synthetic patient, live MedJev answer
STATE
{
 "decision_time": "at emergency department triage",
 "patient": {"age": 67, "sex": "male", "race": "WHITE"},
 "arrival_transport": "AMBULANCE",
 "triage": {"temperature_F": 100.9, "heart_rate_bpm": 118, "respiratory_rate": 24, "o2_saturation_pct": 91,
     "systolic_bp_mmHg": 96, "diastolic_bp_mmHg": 58, "pain": "4", "acuity_esi": 2, "chief_complaint": "Dyspnea,
     fever"},
 "past_year_emergency_department_history": {"visits": 3, "visits_ending_in_admission": 2}
}
QUESTION (Noul): Based on the information available at emergency department triage, will this visit end with the
  patient admitted to the hospital?
  true:  The emergency department disposition is admission to the hospital.
  false: The patient is not admitted (discharged home, transferred, leaves, or another disposition).
GOLD: (synthetic patient, no outcome)
MEDJEV: p(yes) = 0.92
\end{lstlisting}
\end{minipage}

\par\medskip\noindent\begin{minipage}{\linewidth}\textbf{ED Triage Disposition}\par\smallskip
\begin{lstlisting}
[ED Triage Disposition]  synthetic patient, live MedJev answer
STATE
{
 "decision_time": "at emergency department triage",
 "patient": {"age": 67, "sex": "male", "race": "WHITE"},
 "arrival_transport": "AMBULANCE",
 "triage": {"temperature_F": 100.9, "heart_rate_bpm": 118, "respiratory_rate": 24, "o2_saturation_pct": 91,
     "systolic_bp_mmHg": 96, "diastolic_bp_mmHg": 58, "pain": "4", "acuity_esi": 2, "chief_complaint": "Dyspnea,
     fever"},
 "past_year_emergency_department_history": {"visits": 3, "visits_ending_in_admission": 2}
}
QUESTION (Choice): Based on the information available at emergency department triage, what will the disposition
  of this emergency department visit be?
  home  Discharged home from the emergency department.
  admitted  Admitted to the hospital.
  transfer  Transferred to another facility.
  left  Left before the visit was complete (without being seen, eloped, or against medical advice).
  expired  Died in the emergency department.
  other  Another disposition.
GOLD: (synthetic patient, no outcome)
MEDJEV: admitted  (p = 0.94)
  probabilities: home 0.02  admitted 0.94  transfer 0.01  left 0.00  expired 0.01  other 0.02
\end{lstlisting}
\end{minipage}

\par\medskip\noindent\begin{minipage}{\linewidth}\textbf{ED Admission at 2 h}\par\smallskip
\begin{lstlisting}
[ED Admission at 2 h]  synthetic patient, live MedJev answer
STATE
{
 "prediction_time": "2 hours after arrival in the emergency department",
 "patient": {"age": 34, "sex": "female"},
 "past_year_before_this_visit": {"hospital_admissions": 0, "emergency_department_visits": 1, "diagnoses":
     ["Migraine without aura, not intractable"]},
 "this_ed_visit_so_far": {"labs_latest": ["Leukocytes [#/volume] in Blood by Automated count: 7.2 K/uL",
     "Hemoglobin [Mass/volume] in Blood: 13.1 g/dL", "Creatinine [Mass/volume] in Serum or Plasma: 0.7 mg/dL",
     "Glucose [Mass/volume] in Serum or Plasma: 96 mg/dL", "Sodium [Moles/volume] in Serum or Plasma: 139
     mEq/L", "Potassium [Moles/volume] in Serum or Plasma: 4 mEq/L"]}
}
QUESTION (Noul): Based on the patient's record up to now, 2 hours after arriving in the emergency department,
  will this emergency department visit end in admission to the hospital?
  true:  The patient is admitted to the hospital (including observation) from this visit.
  false: The patient leaves the emergency department without being admitted.
GOLD: (synthetic patient, no outcome)
MEDJEV: p(yes) = 0.12
\end{lstlisting}
\end{minipage}

\par\medskip\noindent\begin{minipage}{\linewidth}\textbf{In-Hospital Mortality}\par\smallskip
\begin{lstlisting}
[In-Hospital Mortality]  synthetic patient, live MedJev answer
STATE
{
 "prediction_time": "24 hours after hospital admission",
 "patient": {"age": 81, "sex": "female", "race": "WHITE", "insurance": "Medicare", "language": "ENGLISH",
     "marital_status": "WIDOWED"},
 "current_admission": {"admission_type": "EW EMER.", "admitted_from": "EMERGENCY ROOM", "hours_since_admission":
     24},
 "past_year_before_this_admission": {"hospital_admissions": 2, "diagnoses": ["Chronic diastolic (congestive)
     heart failure", "Chronic kidney disease, stage 4 (severe)", "Unspecified atrial fibrillation"]},
 "this_admission_so_far": {"care_units": ["Emergency Department", "Medical Intensive Care Unit (MICU)"],
     "bedside_measurements_latest": ["Heart rate: 112 bpm", "Non-invasive BP mean: 58 mmHg", "Respiratory rate:
     28 insp/min", "SpO2 (pulse oximetry): 89 %", "GCS verbal response: 3"], "labs_latest": ["Lactate
     [Moles/volume] in Blood: 5.2 mmol/L (high)", "Creatinine [Mass/volume] in Serum or Plasma: 3.9 mg/dL
     (high)", "Leukocytes [#/volume] in Blood by Automated count: 21.4 K/uL (high)", "Bicarbonate [Moles/volume]
     in Serum or Plasma: 14 mEq/L (low)"], "medications": ["Piperacillin-Tazobactam", "Vancomycin", "Heparin"],
     "infusions": ["norepinephrine Injection"]}
}
QUESTION (Noul): Based on the patient's record up to now, will this patient die before being discharged from the
  current hospital admission?
  true:  The patient dies during the current hospital admission.
  false: The patient survives to hospital discharge.
GOLD: (synthetic patient, no outcome)
MEDJEV: p(yes) = 0.86
\end{lstlisting}
\end{minipage}

\par\medskip\noindent\begin{minipage}{\linewidth}\textbf{Length of Stay $>$ 7 Days}\par\smallskip
\begin{lstlisting}
[Length of Stay > 7 Days]  synthetic patient, live MedJev answer
STATE
{
 "prediction_time": "24 hours after hospital admission",
 "patient": {"age": 81, "sex": "female", "race": "WHITE", "insurance": "Medicare", "language": "ENGLISH",
     "marital_status": "WIDOWED"},
 "current_admission": {"admission_type": "EW EMER.", "admitted_from": "EMERGENCY ROOM", "hours_since_admission":
     24},
 "past_year_before_this_admission": {"hospital_admissions": 2, "diagnoses": ["Chronic diastolic (congestive)
     heart failure", "Chronic kidney disease, stage 4 (severe)", "Unspecified atrial fibrillation"]},
 "this_admission_so_far": {"care_units": ["Emergency Department", "Medical Intensive Care Unit (MICU)"],
     "bedside_measurements_latest": ["Heart rate: 112 bpm", "Non-invasive BP mean: 58 mmHg", "Respiratory rate:
     28 insp/min", "SpO2 (pulse oximetry): 89 %", "GCS verbal response: 3"], "labs_latest": ["Lactate
     [Moles/volume] in Blood: 5.2 mmol/L (high)", "Creatinine [Mass/volume] in Serum or Plasma: 3.9 mg/dL
     (high)", "Leukocytes [#/volume] in Blood by Automated count: 21.4 K/uL (high)", "Bicarbonate [Moles/volume]
     in Serum or Plasma: 14 mEq/L (low)"], "medications": ["Piperacillin-Tazobactam", "Vancomycin", "Heparin"],
     "infusions": ["norepinephrine Injection"]}
}
QUESTION (Noul): Based on the patient's record up to now, will the current hospital stay last longer than 7 days
  in total, counted from admission to discharge?
  true:  The hospital stay lasts more than 7 days.
  false: The patient is discharged (or dies) within 7 days of admission.
GOLD: (synthetic patient, no outcome)
MEDJEV: p(yes) = 0.73
\end{lstlisting}
\end{minipage}

\par\medskip\noindent\begin{minipage}{\linewidth}\textbf{30-Day Readmission}\par\smallskip
\begin{lstlisting}
[30-Day Readmission]  synthetic patient, live MedJev answer
STATE
{
 "prediction_time": "at discharge from the current hospital admission",
 "patient": {"age": 72, "sex": "male", "race": "BLACK/AFRICAN AMERICAN", "insurance": "Medicare", "language":
     "ENGLISH", "marital_status": "MARRIED"},
 "current_admission": {"admission_type": "EW EMER.", "admitted_from": "EMERGENCY ROOM", "length_of_stay_days":
     9.5, "discharge_location": "HOME HEALTH CARE"},
 "past_year_before_this_admission": {"hospital_admissions": 3, "diagnoses": ["Chronic systolic (congestive)
     heart failure", "Chronic obstructive pulmonary disease, unspecified"]},
 "this_admission_so_far": {"care_units": ["Emergency Department", "Medicine"], "labs_latest": ["Creatinine
     [Mass/volume] in Serum or Plasma: 1.9 mg/dL", "Sodium [Moles/volume] in Serum or Plasma: 131 mEq/L (low)",
     "Hemoglobin [Mass/volume] in Blood: 10.2 g/dL"], "medications": ["Furosemide", "Metoprolol Succinate XL",
     "Lisinopril", "Insulin"], "diagnoses": ["Acute on chronic systolic (congestive) heart failure", "Chronic
     obstructive pulmonary disease with (acute) exacerbation", "Type 2 diabetes mellitus"], "drg": ["HEART
     FAILURE & SHOCK"]}
}
QUESTION (Noul): Based on the patient's record up to this discharge, will the patient be admitted to the
  hospital again within 30 days after this discharge?
  true:  A new hospital admission starts within 30 days of this discharge.
  false: No new hospital admission within 30 days of this discharge.
GOLD: (synthetic patient, no outcome)
MEDJEV: p(yes) = 0.73
\end{lstlisting}
\end{minipage}

\section{Supervised references}
\label{app:baselines}

Hyperparameters chosen by five-fold cross-validation on the training examples. XGBoost: learning rate 0.05,
subsampling 0.8, a grid over depth $\{2,3,5\}$, minimum child weight $\{1,5\}$ and column subsampling $\{0.3,0.7\}$, with the
number of trees set by early stopping inside each fold. TF-IDF: word $n$-grams $\{(1,1),(1,2)\}$, 50k features, sublinear term
frequency, $\ell_2$ strength $C \in \{0.01, 0.1, 1, 10\}$. For ED Triage Disposition, classes with fewer than 10 training stays
(e.g.\ expired) are merged into ``other'' during training.

\section{Reproducibility}
\label{app:repro}

Everything is produced by the scripts in \texttt{src/scripts/}: \texttt{prepare\_data.py} (Jev requests and training sets), \texttt{extract\_jev.py} and \texttt{run\_variants\_gpu.sh} (zero-shot answers for every
language model; MLX), \texttt{train\_probes.py} (recalibration), \texttt{run\_baselines.py} (supervised references),
\texttt{label\_efficiency.py} (label budgets, from cached outputs), \texttt{evaluate.py} (scores and metrics),
\texttt{make\_paper\_assets.py} (tables and every number in this paper) and \texttt{make\_figures.py} (figures). Every
model output is cached on disk, keyed by the exact request.

\end{document}